\documentclass[times,review,10pt]{elsarticle}

\usepackage{amssymb}
\usepackage{amsmath}
\usepackage{multirow}
\usepackage{graphicx}
\usepackage{wrapfig}
\usepackage{xcolor}

\journal{Pattern Recognition}

\begin{document}

\begin{frontmatter}



\title{Towards Arbitrary-Scale Spacecraft Image Super-Resolution via Salient Region-Guidance}


\author[a]{Jingfan Yang} 
\author[a]{Hu Gao} 
\author[a]{Ying Zhang} 
\author[a]{Depeng Dang\corref{cor1}} 

\affiliation[a]{organization={School of Artificial Intelligence, Beijing Normal University},
            addressline={No.19, Xinjiekouwai St.}, 
            city={Beijing},
            postcode={100875}, 
            country={China}}

\cortext[cor1]{Corresponding author: Depeng Dang (ddepeng@bnu.edu.cn)}

\begin{abstract}
Spacecraft image super-resolution seeks to enhance low-resolution spacecraft images into high-resolution ones. Although existing arbitrary-scale super-resolution methods perform well on general images, they tend to overlook the difference in features between the spacecraft core region and the large black space background, introducing irrelevant noise. In this paper, we propose an efficient salient region-guided spacecraft image arbitrary-scale super-resolution network (SGSASR), which uses features from the spacecraft core salient regions to guide latent modulation and achieve arbitrary-scale super-resolution. Specifically, we design a spacecraft core region recognition block (SCRRB) that identifies the core salient regions in spacecraft images using a pre-trained saliency detection model. Furthermore, we present an adaptive-weighted feature fusion enhancement mechanism (AFFEM) to selectively aggregate the spacecraft core region features with general image features by dynamic weight parameter to enhance the response of the core salient regions. Experimental results on spacecraft radar image dataset and optical image dataset demonstrate that the proposed SGSASR outperforms state-of-the-art approaches. The codes are available at: https://github.com/shenduke/SGSASR.
\end{abstract}




\end{frontmatter}



\section{Introduction}
Spacecraft image is the on-orbit spacecraft target image obtained through radar or optical technology. The clear and high-quality spacecraft image is the fundamental assurance for successfully completing many space missions such as spacecraft pose estimation~\cite{yang2020hcnn, zhang2024monocular}, spacecraft control~\cite{darban2025carla, lopez2024applying} and spacecraft 3D reconstruction~\cite{liu2021three, long20243}. However, due to the large imaging distance, the harsh space lighting environment and sensor interference, it is difficult to obtain high-resolution (HR) spacecraft image~\cite{murray2021mask}.

Spacecraft image super-resolution (SR) aims to restore HR spacecraft images from low-resolution (LR) spacecraft images. Although many spacecraft image SR methods have been proposed~\cite{yang2021srdn, yang2023ssrn}, they typically learn upsampling functions for a specific scale factor and require training different models for different SR scales~\cite{cao2023ciaosr}. When there is a lack of suitable scale training data or the degradation scale of images is unfixed, these methods yield unsatisfactory performance. In practical aerospace application scenarios, it is relatively difficult to obtain real spacecraft HR images to construct training datasets. The degradation scale of different images in the same spacecraft image dataset is also unfixed because of the complex space environmental conditions. In addition, storing a large number of models with different super-resolution scales simultaneously will bring enormous storage and computational pressure to spacecraft equipment.

In contrast, arbitrary-scale SR uses a single model to process LR images at arbitrary scales, which is more suitable for practical image SR scenarios. Most current arbitrary-scale SR methods, such as LIIF~\cite{chen2021learning} and LTE~\cite{lee2022local}, are based on implicit neural representation (INR), and use the implicit neural function to replace the upsampling module in traditional image SR methods. However, these methods are aimed at general images and failed to fully consider the characteristics of spacecraft images.

\begin{figure}[t]
\centering
\includegraphics[width=1.0\linewidth]{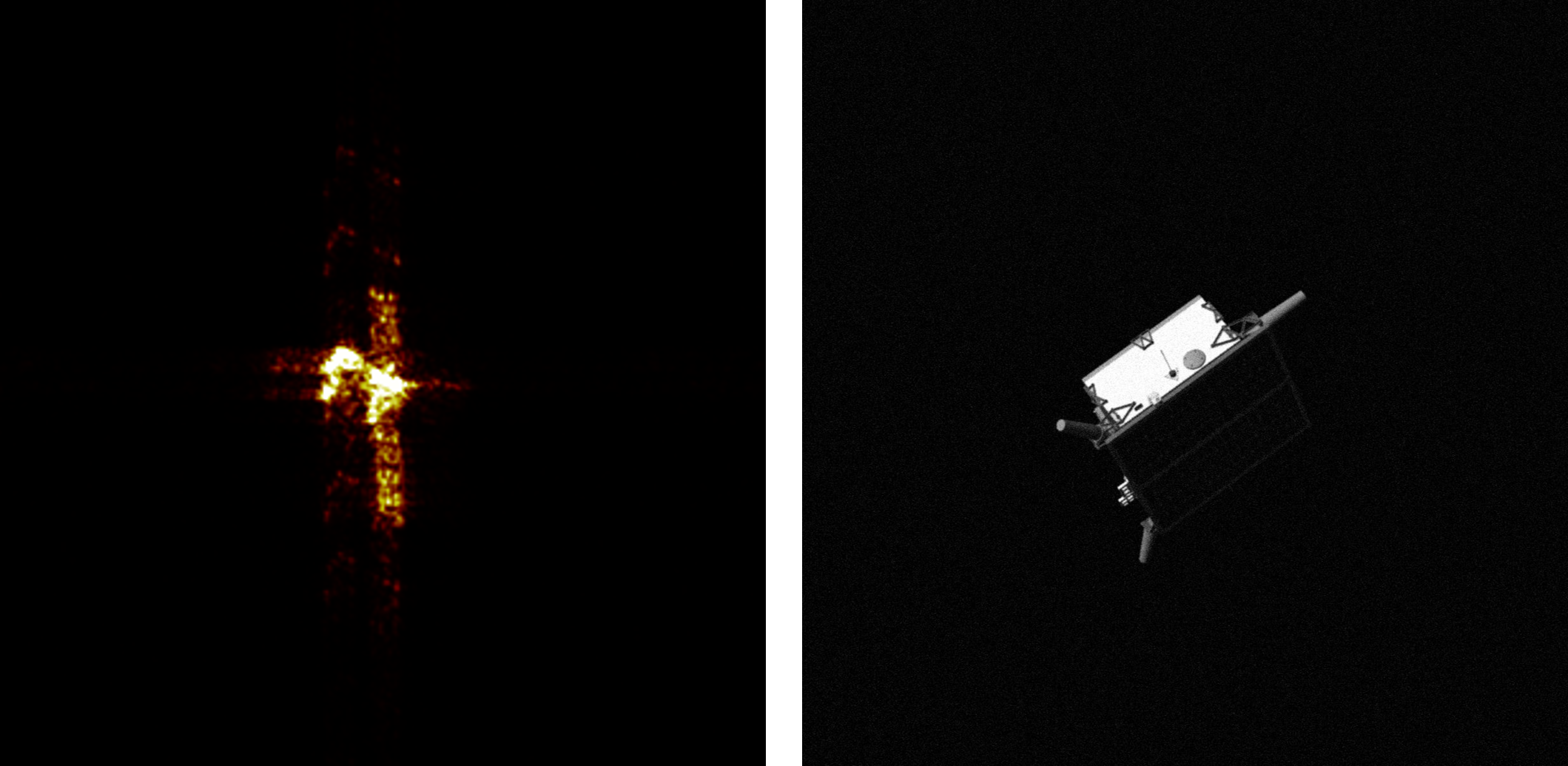}
\caption{Example of spacecraft image, with a large black space background.}
\label{fig:example}
\end{figure}

As shown in Fig. \ref{fig:example}, spacecraft images are acquired from the outer space environment, and mostly with a large black space background. The existing arbitrary-scale SR methods regard each region in spacecraft images equally, overlooking the differences in features between the spacecraft core region and the background region~\cite{liu2021saliency}. As a result, these models are prone to falling into local suboptimal solutions or even introducing noise to the black background region. This limitation makes it difficult for existing arbitrary-scale SR methods to be effectively applied to spacecraft image SR tasks. Therefore, how to better focus on restoring the spacecraft core region is the key to improving the performance of the spacecraft image SR task.

In addition, spacecraft image application scenarios, such as spacecraft control and autonomous navigation, require high real-time performance to ensure timely decision-making and reliable execution of critical operations.

To address these challenges, we propose an efficient salient region-guided spacecraft image arbitrary-scale super-resolution network (SGSASR) in this paper. Specifically, we design an arbitrary-scale image SR framework based on latent modulated function (LMF), which generates latent modulation guided by the spacecraft core salient region, and then decodes the latent modulations to generate spacecraft SR images of different scales according to requirements. Furthermore, we design a spacecraft core region recognition block (SCRRB) to identify the core salient region in spacecraft LR images, allowing the network to focus on restoring the region. Finally, we propose an adaptive-weighted feature fusion enhancement mechanism (AFFEM), which selectively aggregates the spacecraft core region features with general image features to enhance the response of the core salient region.

Our main contributions are summarized as follows: 
\begin{itemize}
\item We propose the first method to achieve spacecraft image arbitrary-scale SR, called SGSASR, which can super-resolve spacecraft LR images at arbitrary scales through a single model.
\item We propose a spacecraft core region recognition block (SCRRB) that employs saliency detection to identify the core salient region in spacecraft images.
\item We propose an adaptive-weighted feature fusion enhancement mechanism (AFFEM) to enhance the response of the spacecraft core salient region in feature maps.
\item Extensive experiments demonstrate that our SGSASR not only achieves state-of-the-art performance on spacecraft image arbitrary-scale SR while maintaining high computational efficiency, but also achieves competitive performance compared to fixed-scale SR methods.
\end{itemize}

This paper is organized as follows. Section \ref{sec:related_work} introduces the related works. Section \ref{sec:method} describes our proposed method in detail. Section \ref{sec:experiment} presents the experimental results and analysis. Section \ref{sec:conclusion} concludes the paper.

\section{Related Work}
\label{sec:related_work}
\subsection{Single Image Super-Resolution}
Single Image Super-Resolution (SISR) is a classic image SR task, and a large number of deep learning-based SISR methods have been proposed in recent years~\cite{zhou2023srformer, chen2024single}. These methods can be classified into two main categories: CNN-based and Transformer-based.

CNN is the first type of deep learning method introduced into SISR. SRCNN~\cite{dong2014learning} is the first deep learning-based SISR method, which represents the end-to-end mapping between LR/HR images as CNN. Lim et al.~\cite{lim2017enhanced} remove unnecessary modules from traditional residual networks and propose an enhanced deep super-resolution network (EDSR). Zhang et al.~\cite{zhang2018image} propose the residual channel attention network (RCAN), which leverages a very deep residual-in-residual structure and a channel attention mechanism to enhance the learning of high-frequency details in image SR. Wang et al.~\cite{wang2018esrgan} introduce GAN into SISR and utilize the residual-in-residual dense block, the relativistic discriminator, and the refined perceptual loss to achieve better texture restoration and visual quality. Zhou et al.~\cite{zhou2022efficient} propose VapSR that evolves a simple pixel attention module by expanding the receptive field, adopting depthwise separable convolutions and pixel normalization. Gao et al.~\cite{gao2025learning} introduce a lattice structure and significantly improve the computational efficiency of image SR by utilizing lattice stereo NAFBlock and the lattice stereo attention module. Chen et al.~\cite{chen2024large} introduce the large kernel frequency-enhanced network (LKFN), which uses a novel frequency-enhanced pixel attention mechanism to overcome the limitations of spatial-domain attention in lightweight image SR.

Transformer has also been introduced into SISR in recent years. IPT~\cite{chen2021pre} is the first Transformer-based SISR method, which adapts to different image processing tasks by pre-training on the ImageNet benchmark. Liang et al.~\cite{liang2021swinir} propose SwinIR, an image SR method based on the Swin Transformer, which combines shallow feature extraction, deep feature extraction with residual Swin Transformer blocks, and high-quality image reconstruction. Zhang et al.~\cite{zhang2022efficient} introduce an efficient long-range attention network (ELAN) that combines shift convolution and a group-wise multi-scale self-attention module to reduce computational complexity while exploiting long-range image dependencies. Zhao et al.~\cite{zhao2023spherical} propose a spherical space feature decomposition network that effectively separates and aligns modality-specific features by contrastive learning in the spherical space. Xie et al.~\cite{xie2025mat} propose a Multi-Range Attention Transformer (MAT), which flexibly integrates attention across diverse spatial extents using dilation-based operations. Li et al.~\cite{li2025srconvnet} propose a lightweight SISR framework (SRConvNet) that integrates the global modeling strengths of transformers with the efficiency of CNNs by introducing Fourier-modulated attention and dynamic mixing layer. Xiao et al.~\cite{xiao2024ttst} creatively introduce the top-k token selective mechanism, dynamically selecting the top-k keys in terms of score ranking for each query to obtain the most crucial token.

\subsection{Spacecraft Image Super-Resolution}
The spacecraft image super-resolution (SR) task aims to generate high-resolution (HR) spacecraft images from low-resolution (LR) inputs. The early spacecraft image SR methods typically required manually designed prior features~\cite{bulyshev2010computational, murthy2014skysat}. In recent years, deep learning-based methods have gradually been introduced into the spacecraft image SR task. Zhang et al.~\cite{zhang2019comparable} propose a spacecraft image SR method based on the deep recursive convolutional network and fine-tune it on the simulated space object dataset. Yang et al.~\cite{yang2021srdn} propose a unified GAN method for spacecraft image SR and deblurring, which restores texture information by designing symmetric downsampling and upsampling modules. Shi et al.~\cite{shi2023super} introduce the dual regression and deformable convolutional attention mechanism to better extract the high-frequency features of spacecraft LR images. Yang et al.~\cite{yang2023ssrn} propose a spacecraft ISAR image SR method based on non-local sparse attention, which can capture long-range self-similarity in spacecraft images. Li et al.~\cite{li2023nlsan} present a non-local scene awareness network (NLSAN), which incorporates a non-local interpolation approach to capture global textures in spacecraft images. Gao et al.~\cite{gao2024learning} propose a sparse attention and selective cross-fusion network (SSNet) for spacecraft ISAR image SR, which introduces the top-k selection operator to enhance the distinction between background and spacecraft information.

\subsection{Arbitrary-Scale Super-Resolution}
Arbitrary-scale super-resolution is a task that uses a single model to achieve SR at any scale, so it is more practical. Hu et al.~\cite{hu2019meta} propose MetaSR, which uses meta-module to dynamically predict the weights of large-scale filters based on input scale. Chen et al.~\cite{chen2021learning} propose the local implicit image function (LIIF) inspired by implicit neural representation (INR), which predicts RGB values at arbitrary coordinates using local deep features. Subsequently, INR becomes the main paradigm for arbitrary-scale SR tasks~\cite{yao2023local, hu2025gaussiansr}. Lee et al.~\cite{lee2022local} enhance implicit neural functions by capturing fine image textures in the 2D Fourier space. He et al.~\cite{he2024latent} introduce a computationally optimal paradigm for continuous image representation in arbitrary-scale SR by decoupling HR decoding into a shared latent space and a lightweight rendering space. Zhao et al.~\cite{zhao2024activating} propose SLFAM and LFAUM that enable fixed-scale SR networks to effectively handle arbitrary scales by incorporating both scale factors and local feature diversity. Wei et al.~\cite{wei2025multi} propose a Dynamic Implicit Network (DINet) for multi-contrast MRI arbitrary-scale SR, which uses the scale-adaptive dynamic convolution to learn multi-contrast feature representations. Zhu et al.~\cite{zhu2025multi} propose the multi-scale implicit transformer (MSIT), which introduces multi-scale characteristics into arbitrary-scale SR for the first time, enhancing latent code adaptability across magnification factors. Zhang et al.~\cite{zhang2025joint} propose DASU-Net and its fast version FDASU-Net, which is a joint depth map super-resolution and high–low frequency decomposition framework capable of arbitrary-scale up-sampling and robust cross-modal feature fusion.

\section{Methodology}
\label{sec:method}


\begin{figure}[t]
\centering
\includegraphics[width=1.0\linewidth]{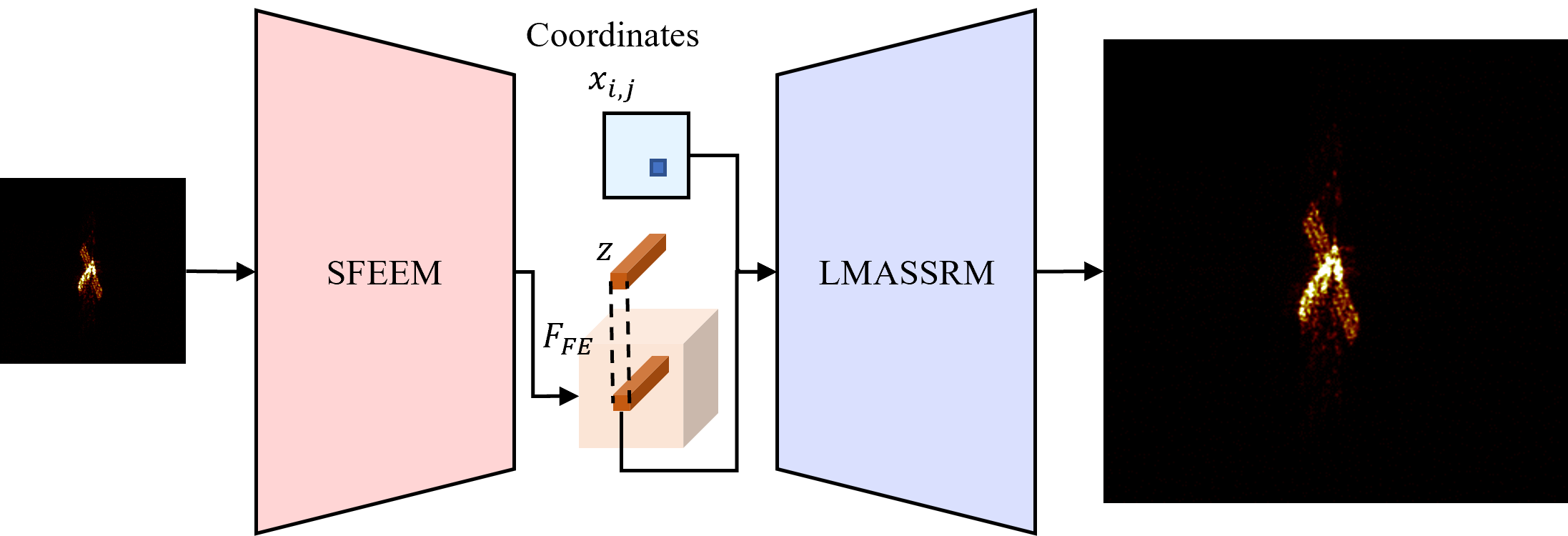}
\caption{The overall architecture of salient region-guided spacecraft image arbitrary-scale super-resolution network (SGASR).}
\label{fig:SGSASR}
\end{figure}

\subsection{Overall Architecture}
The overall architecture of our SGSASR is shown in Fig. \ref{fig:SGSASR}, which can be divided into two main components: the spacecraft image feature extraction enhancement module (SFEEM) as the encoder and the latent modulation-based arbitrary-scale super-resolution module (LMASSRM) as the decoder. 

We first input the spacecraft LR image $I_{LR} \in \mathbb{R}^{h \times w \times C_{in}}$ into SFEEM to obtain spacecraft core region enhanced features $F_{FE} \in \mathbb{R}^{h \times w \times D_{FE}}$.
\begin{eqnarray}
    F_{FE} = SFEEM(I_{LR})
\end{eqnarray}%
where $SFEEM(\cdot)$ denotes the spacecraft image feature extraction enhancement module.

Subsequently, we derive a feature vector from $F_{FE}$, denoted as the latent code $z \in \mathbb{R}^{D_{FE}}$, which is used to decode the nearest coordinates in the continuous spatial domain $X \in \mathbb{R}^2$.

Finally, we input $z$ and the nearest coordinate $x_{i,j} \in X$ into the LMASSRM to decode the pixel values at each HR coordinate, ultimately achieving the arbitrary-scale spacecraft SR images.
\begin{eqnarray}
    I_{SR}(x_{i,j}) = LMASSRM(z,x_{i,j})
\end{eqnarray}%
where $I_{SR}(x_{i,j})$ is the predicted pixel value at $x_{i,j}$ of arbitrary-scale spacecraft SR image $I_{SR} \in \mathbb{R}^{H \times W \times C_{out}}$, $LMASSRM(\cdot)$ denotes the latent modulation-based arbitrary-scale super-resolution module.

We optimize SGSASR using L1 loss:
\begin{eqnarray}
\mathcal{L}_{L_1} = \| I_{HR} - I_{SR} \|_{1}
\end{eqnarray}%
where $I_{HR}$ is the ground-truth spacecraft HR image.

\begin{figure*}[t]
\centering
\includegraphics[width=1.0\linewidth]{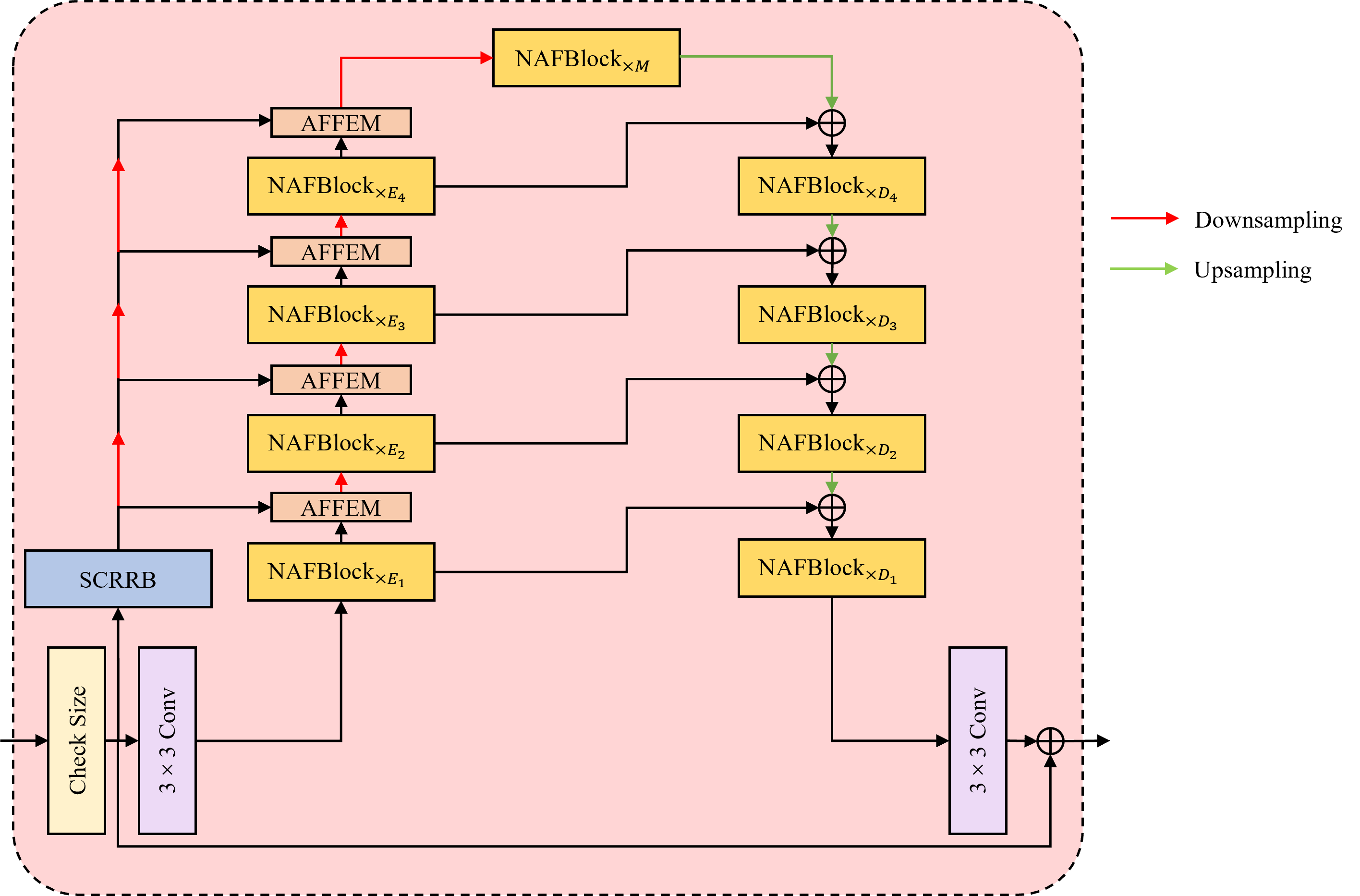}
\caption{The architecture of spacecraft image feature extraction enhancement module (SFEEM). }
\label{fig:SFEEM}
\end{figure*}

\subsection{Spacecraft Image Feature Extraction Enhancement Module (SFEEM)}
Although existing arbitrary-scale SR methods for general images achieve well performance, they often overlook the differences in features between the spacecraft core region and the surrounding large black background region, introducing irrelevant noise to the black background region. To address this problem, we propose the spacecraft image feature extraction enhancement module (SFEEM), which identifies the spacecraft salient core region within the image and enhances its response. Focusing on the restoration of the spacecraft core region is essential to improve the performance of spacecraft image SR.

The structure of SFEEM is shown in Fig. \ref{fig:SFEEM} (a). For the input spacecraft LR image $I_{LR}$, we first use the spacecraft core region recognition block to identify the spacecraft salient core region in the image, while utilizing a $3 \times 3$ convolution layer to extract shallow feature maps. Next, these shallow feature maps are processed by an encoder sub-network consisting of several NAFBlocks, gradually reducing the size of the feature maps while expanding the channels. Among them, the input of each downsampling layer is an enhanced feature map obtained by fusing the spacecraft core region features with the previous NAFBlock output features by adaptive-weighted feature fusion enhancement mechanism. Then, the deep feature maps output by the encoder are restored to the original size by the middle block and decoder. To ensure the stability of training, we add the skip connection. Finally, we use a $3 \times 3$ convolution to generate residual feature maps for subsequent module processing.

\begin{figure*}[t]
\centering
\includegraphics[width=1.0\linewidth]{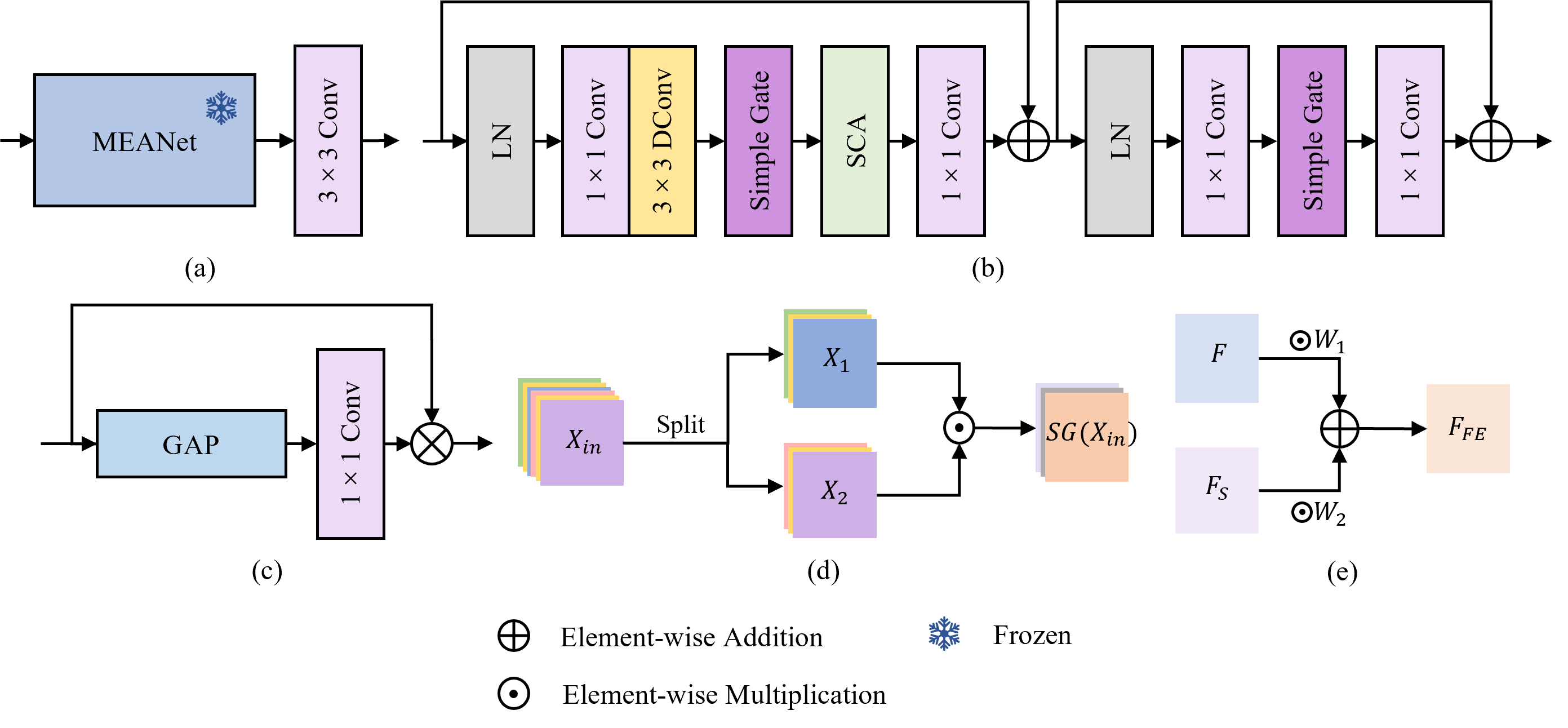}
\caption{(a) Spacecraft core region recognition block (SCRRB). (b) Nonlinear activation free block (NAFBlock). (c) Simplified channel attention (SCA). (d) Simple gate. (e) Adaptive-weighted feature fusion enhancement mechanism (AFFEM).}
\label{fig:SFEEM_detail}
\end{figure*}

\subsubsection{Spacecraft Core Region Recognition Block (SCRRB)}
In order to identify the spacecraft core region in the image, we need to perform saliency object detection on the input spacecraft image. However, directly inserting a saliency detection module into the network structure cannot guarantee detection accuracy and increases the additional computational burden. Therefore, we propose to introduce a pre-trained object detection model and construct SCRRB based on it. The architecture of SCRRB is shown in Fig. \ref{fig:SFEEM_detail} (a).

We first introduce a lightweight salient object detection method MEANet~\cite{liang2024meanet} as the core component of our SCRRB to preliminarily detect the target saliency of the input spacecraft LR image $I_{LR}$. 
\begin{eqnarray}
    S = MEANet^{*}(I_{LR})
\end{eqnarray}%
where $S$ is the predicted map of the identified core salient region, $MEANet^{*}(\cdot)$ is the pre-trained MEANet and has been frozen. This approach not only utilizes the prior knowledge advantage of MEANet, but also reduces the computational burden of local retraining.

Then, we use a convolution layer to extract the feature map from the predicted map of the core salient region.
\begin{eqnarray}
    F_{S} = f^{c}_{3 \times 3}(S)
\end{eqnarray}%
where $F_{S}$ is the feature map that contains the recognition results of the core salient region, and $f^{c}_{3 \times 3}(\cdot)$ denotes the $3 \times 3$ convolution layer.

\subsubsection{Nonlinear Activation Free Block (NAFBlock)}
In order to improve computational efficiency, we utilize NAFBlock~\cite{chen2022simple} as the base block of the encoder-decoder in SFEEM. NAFBlock introduces the simple gate function to replace the nonlinear activation function, as shown in Fig. \ref{fig:SFEEM_detail} (b). Formally, the detailed process of NAFBlock is represented as follows:
\begin{align}
    X_{1} &= X_{in} + f^{c}_{1 \times 1}(SCA(SG(f^{dwc}_{3 \times 3}(f^{c}_{1 \times 1}(LN(X_{in})))))) \\
    X_{F} &= X_{1} + f^{c}_{1 \times 1}(SG(f^{c}_{1 \times 1}(LN(X_{1}))))
\end{align}%
where $X_{in}$ and $X_{F}$ are respectively the input and output feature maps of NAFBlock, $X_{1}$ are the intermediate feature map, $LN(\cdot)$ denotes the layer normalization,  $f^{c}_{1 \times 1}(\cdot)$ and $f^{dwc}_{3 \times 3}(\cdot)$ respectively denote the $1 \times 1$ convolution and the $3 \times 3$ depthwise convolution, $SG(\cdot)$ and $SCA(\cdot)$ respectively denote simple gate and simplified channel attention (SCA).

In SCA, the unnecessary convolution layers and nonlinear activation functions have been removed.
\begin{eqnarray}
    SCA &= X_{f_{3}} \odot f^{c}_{1 \times 1}(GAP(X_{f_{3}}))
\end{eqnarray}%
where $GAP(\cdot)$ denotes the global average pooling.

In simple gate, the feature map is divided directly into two parts in the channel dimension and then multiplied together.
\begin{eqnarray}
    SG = X_{f_{1}} \odot X_{f_{2}} 
\end{eqnarray}%
where $X_{f_{1}}$ and $X_{f_{2}}$ are the two equal parts of the input feature map that are divided, $\odot$ denotes the element-wise multiplication.

\subsubsection{Adaptive-Weighted Feature Fusion Enhancement Mechanism (AFFEM)}
The traditional feature fusion methods such as summation and concatenation lack flexible expression, which limits the performance of feature fusion. Therefore, we propose AFFEM as shown in Fig. \ref{fig:SFEEM_detail} (e), which achieves adaptive feature fusion enhancement by dynamic weight parameters. Formally, the process of our AFFEM can be defined as:
\begin{eqnarray}
    F_{FE} = W_1 \odot F + W_2 \odot F_{S}
\end{eqnarray}%
where $F$ is the features from the previous NAFBlock, $F_{FE}$ is the enhanced features after fusion, $W_1$ and $W_2$ both denote the learnable parameters, directly optimized by backpropagation and initialized as 1.

\begin{figure*}[t]
\centering
\includegraphics[width=0.8\linewidth]{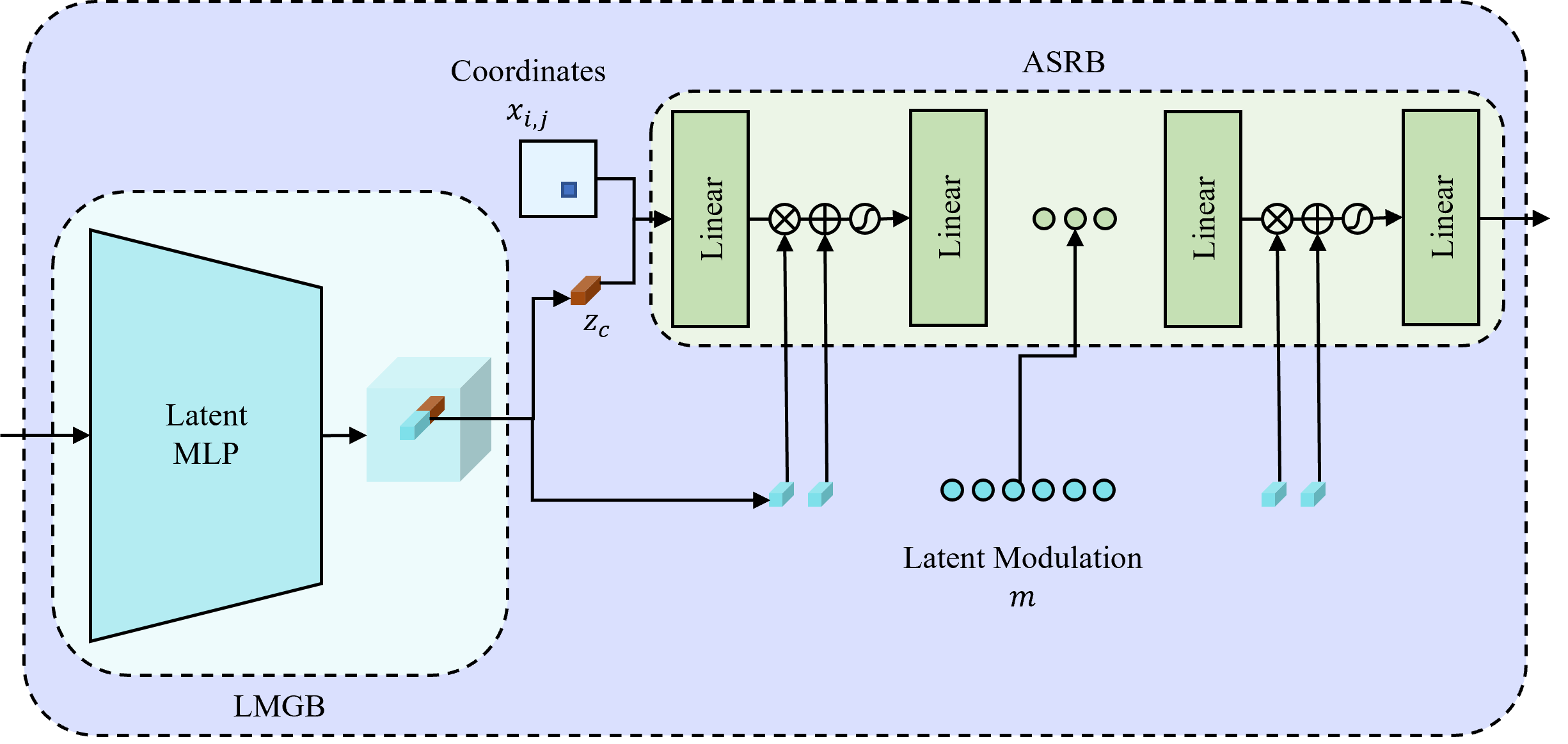}
\caption{The architecture of latent modulation-based arbitrary-scale super-resolution module (LMASSRM).}
\label{fig:LMASSRM}
\end{figure*}

\subsection{Latent Modulation-Based arbitrary-scale super-resolution Module (LMASSRM)}
In order to achieve arbitrary-scale spacecraft SR images, we propose a two-stage latent modulation-based arbitrary-scale super-resolution module (LMASSRM). The architecture of LMASSRM is shown in Fig. \ref{fig:LMASSRM}. In the first stage, we use the latent modulation generation block (LMGB) to generate the latent modulation and compress the latent code. In the second stage, we use the arbitrary-scale restoration block (ASRB) to dynamically adjust the hidden features according to the latent modulation signals from the previous stage, thereby generating pixel values at arbitrary scales. Formally, the detailed process of LMASSRM is represented as follows:
\begin{align}
    [m,z_{c}] &= LMGB(z) \\
    I(x_{i,j}) &= ASRB_{m(z_{c},x_{i,j})}
\end{align}%
where $z$ is the latent code, which is the feature vector derived from the enhanced features $F_{FE}$, used to decode its nearest coordinates in the continuous spatial domain, $z_{c}$ is the compressed latent code, $m \in \mathbb{R}^{D_m}$ is the latent modulation, $x_{i,j}$ denotes any nearest coordinate of $z$, $I(x_{i,j})$ denotes the predicted pixel value at $x_{i,j}$, $LMGB(\cdot)$ denotes latent modulation generation block, $ASRB_{m}(\cdot)$ denotes arbitrary-scale restoration block which needs to dynamically adjust the hidden features according to the latent modulation signals.

\subsubsection{Latent Modulation Generation Block (LMGB)}
In LMGB, we use latent MLP to generate latent modulation $m$ according to the input latent code $z$, and compress the high-dimensional latent code $z$ into the low-dimensional latent code $z_{c}$. The linear layers in the latent MLP are modulated using the FiLM mechanism~\cite{perez2018film}, which includes the scale modulation and the shift modulation. The latent modulation $m$ encodes the input features and provides dynamic parameters for subsequent restoration.
\begin{eqnarray}
   m = [\alpha_{1},\beta_{1},\alpha_{2},\beta_{2},...,\alpha_{K},\beta_{K}] = Split(MLP_{l}(z)) 
\end{eqnarray}%
where $\alpha$ denotes the element-wise scale modulation, $\beta$ denotes the element-wise shift modulation, $K$ denotes the depth of hidden layers in subsequent restoration, $MLP_{l}(\cdot)$ denotes the latent MLP.

\subsubsection{Arbitrary Scale Restoration Block (ASRB)}
In ASRB module, we input the low-dimensional latent code $z_{c}$ and the nearest coordinates $(x,y)$ into the render MLP to decode and predict the pixel values at the HR coordinates at any resolution. The render MLP consists of 7 layers, and the features in each layer are dynamically adjusted based on the latent modulation $m$ of the latent code $z_{c}$ to achieve fast matching. The detailed process of dynamic adjustment is represented as follows:
\begin{eqnarray}
   h_{k+1} = MLP_{r^{k+1}}(\sigma((1+\alpha_{k})\odot h_{k}+\beta_{k}))
\end{eqnarray}%
where $h_{k}$ and $h_{k+1}$ respectively denote the features after the $k$-th render MLP layer and the $k+1$-th render MLP layer, $\sigma$ denotes the $ReLU$ activation function.

\section{Experiment}
\label{sec:experiment}

In this section, we first introduce the experimental settings, then perform quantitative and qualitative comparisons with existing methods, and finally conduct ablation studies to validate the effectiveness of our approach. 

\subsection{Experimental Settings}

\subsubsection{Datasets}
In this paper, we conduct experiments on two spacecraft image datasets. The first is a real spacecraft radar image dataset, and the second is the publicly available spacecraft optical image dataset SPEED~\cite{sharma2020neural}.

The real spacecraft radar image dataset consists of real inverse synthetic aperture radar (ISAR) images collected during the in-orbit flight of a certain type of non-cooperative spacecraft. It contains a total of 5,250 images, with 4,800 used for training and 450 for evaluation.

SPEED is the PRISMA Tango satellite image dataset, originally used for the spacecraft pose estimation challenge organized by European Space Agency (ESA) and Stanford University’s Space Rendezvous Lab (SLAB). We use 7,499 images with a black space background, among which 6,000 are used for training and 1,499 for evaluation.

\subsubsection{Evaluation Metrics}
We use two popular image quality evaluation metrics: peak signal-to-noise ratio (PSNR) and structural similarity (SSIM). The higher values of these two metrics indicate the better performance of the method. Furthermore, we use Floating Point Operations (FLOPs) to evaluate the computational complexity, the lower value of the FLOPs indicates the lower the computational complexity and the higher the computing efficiency of the method. 

\subsubsection{Implementation Details}
We first introduce the configuration of each module in SGSASR. In SFEEM, we apply [2,2,4,8] NAFBlock at each scale of the encoder sub-network, [2,2,2,2] NAFBlock at each scale of the decoder sub-network, and the number of NAFBlocks in the middle block is 12. In LMGB, we set the input dimension of the latent MLP to 128 and the output dimension to 288. In ASRB, we set the feature dimension for the render MLP to 16.

Then we introduce the specific training details. We use NVIDIA GeForce RTX 3090 GPU to train the SGSASR. We train models with Adam optimizer and L1 loss for 200 epochs with the initial learning rate 2e-4, and decayed by a factor of 0.5 at [50, 100, 150, 175] epochs. We set the batch size to 16. The patch size is set to 48 for spacecraft ISAR image and 128 for spacecraft optical image.

\begin{table*}[t]
\caption{Quantitative comparisons with other fixed-scale SR methods on spacecraft radar image dataset. The best and the second best results are marked in \pmb{bold} and \underline{underline}.}
\label{tab:isar_fix_results}
\centering
\begin{tabular}{lllll}
\hline
\multirow{2}{*}{Methods}& \multicolumn{2}{c}{$\times$4}  & \multicolumn{2}{c}{$\times$8} \\ 
  & \multicolumn{1}{c}{PSNR$\uparrow$}  & \multicolumn{1}{c}{SSIM$\uparrow$} & \multicolumn{1}{c}{PSNR$\uparrow$}  & \multicolumn{1}{c}{SSIM$\uparrow$} \\
\hline
SSRN~\cite{yang2023ssrn} & \underline{32.93} & \pmb{0.9847} & \underline{28.64} & \pmb{0.9730} \\
SSNet~\cite{gao2024learning} & 32.23 & \underline{0.9828} & 27.91 & 0.9627 \\
VapSR~\cite{zhou2022efficient} & 32.32 & 0.9825 & 28.30 & 0.9667 \\
LKFN~\cite{chen2024large} & 31.77 & 0.9817 & 27.94 & 0.9640 \\
ELAN~\cite{zhang2022efficient} & 31.06 & 0.9678 & 28.06 & 0.9340  \\
MAT~\cite{xie2025mat} & 32.07 & 0.9781 & 28.56 & 0.9568 \\
SRConvNet~\cite{li2025srconvnet} & 32.37 & 0.9822 & 28.17 & 0.9634 \\
SGSASR(Ours) & \pmb{34.07} & 0.9823 & \pmb{29.03} & \underline{0.9723}\\
\hline
\end{tabular}
\end{table*}

\begin{table*}[t]
\caption{Quantitative comparisons with other arbitrary-scale SR methods on spacecraft radar image dataset. The best and the second best results are marked in \pmb{bold} and \underline{underline}.}
\label{tab:isar_assr_results}
\centering
\resizebox{\textwidth}{!}{
\begin{tabular}{llllllllll}
\hline
\multirow{2}{*}{Methods}& \multicolumn{2}{c}{$\times$4}  & \multicolumn{2}{c}{$\times$6} & \multicolumn{2}{c}{$\times$8} & \multicolumn{2}{c}{$\times$12} & \multirow{2}{*}{FLOPs$\downarrow$}\\ 
  & \multicolumn{1}{c}{PSNR$\uparrow$}  & \multicolumn{1}{c}{SSIM$\uparrow$} & \multicolumn{1}{c}{PSNR$\uparrow$}  & \multicolumn{1}{c}{SSIM$\uparrow$} & \multicolumn{1}{c}{PSNR$\uparrow$}  & \multicolumn{1}{c}{SSIM$\uparrow$} & \multicolumn{1}{c}{PSNR$\uparrow$}  & \multicolumn{1}{c}{SSIM$\uparrow$} \\
\hline
LIIF~\cite{chen2021learning} & 33.51 & \pmb{0.9826} & 30.15 & \pmb{0.9765} & 28.95 & \pmb{0.9730} & \underline{28.05} & \pmb{0.9707} & 160.64G \\
LM-LIIF~\cite{he2024latent}  & \pmb{34.17} & \underline{0.9825} & \underline{30.19} & 0.9757 & \underline{28.99} & 0.9721 & 27.99 & 0.9699 & 20.99G \\
LM-LTE~\cite{he2024latent} & 33.03 & 0.9819 & 30.07 & 0.9755 & 28.88 & 0.9718 & 28.04 & 0.9685 & 21.55G\\
SGSASR(Ours) & \underline{34.07} & 0.9823 & \pmb{30.24} & \underline{0.9758} & \pmb{29.03} & \underline{0.9723} & \pmb{28.22} & \underline{0.9700} & \pmb{4.37G} \\
\hline
\end{tabular}
}
\end{table*}

\begin{figure*}[t]
\centering
\begin{minipage}[t]{0.45\linewidth}
  \centering
  \includegraphics[width=1.0\textwidth]{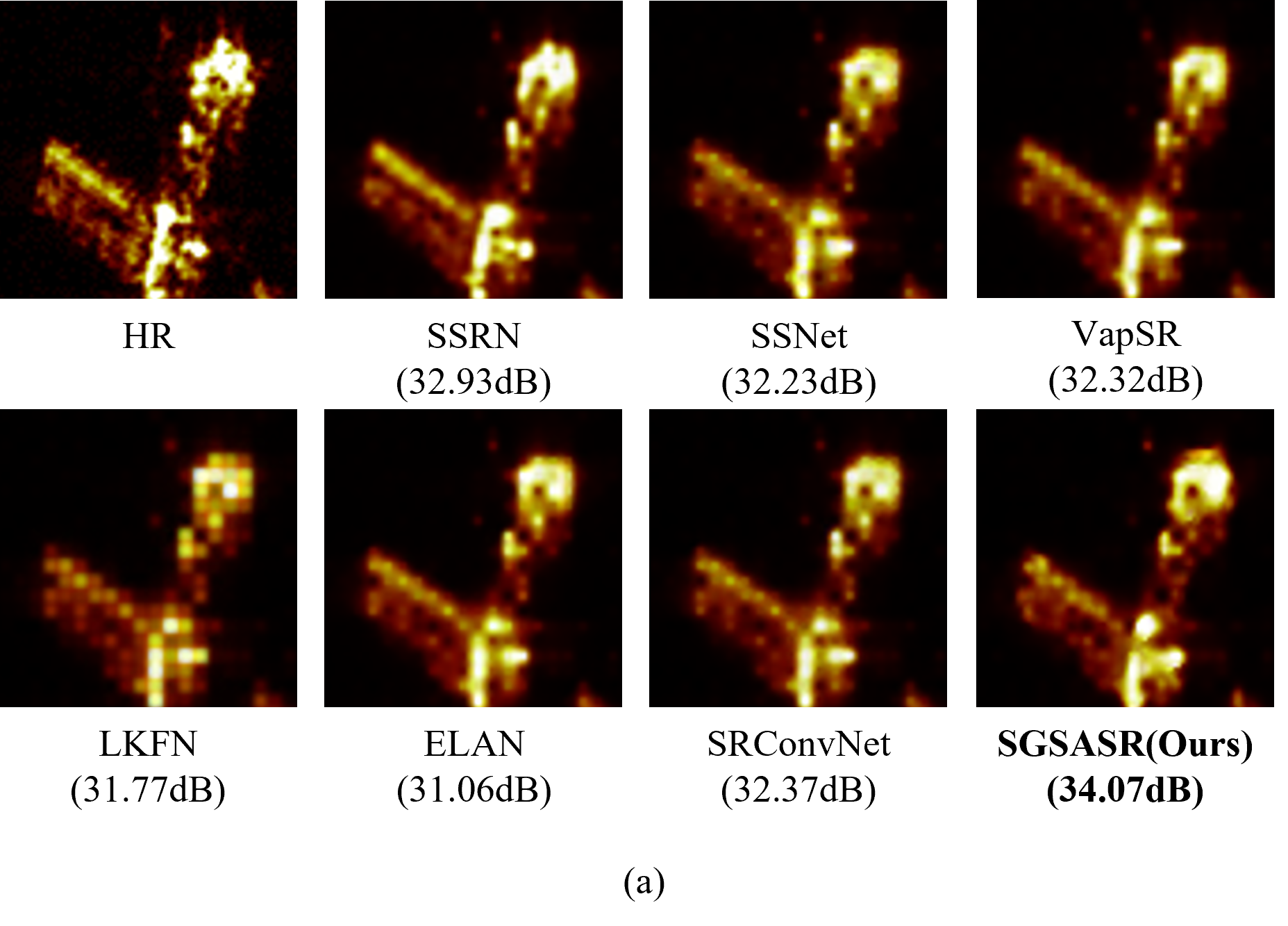}
\end{minipage}  
\quad
\begin{minipage}[t]{0.45\linewidth}
  \centering
  \includegraphics[width=1.0\textwidth]{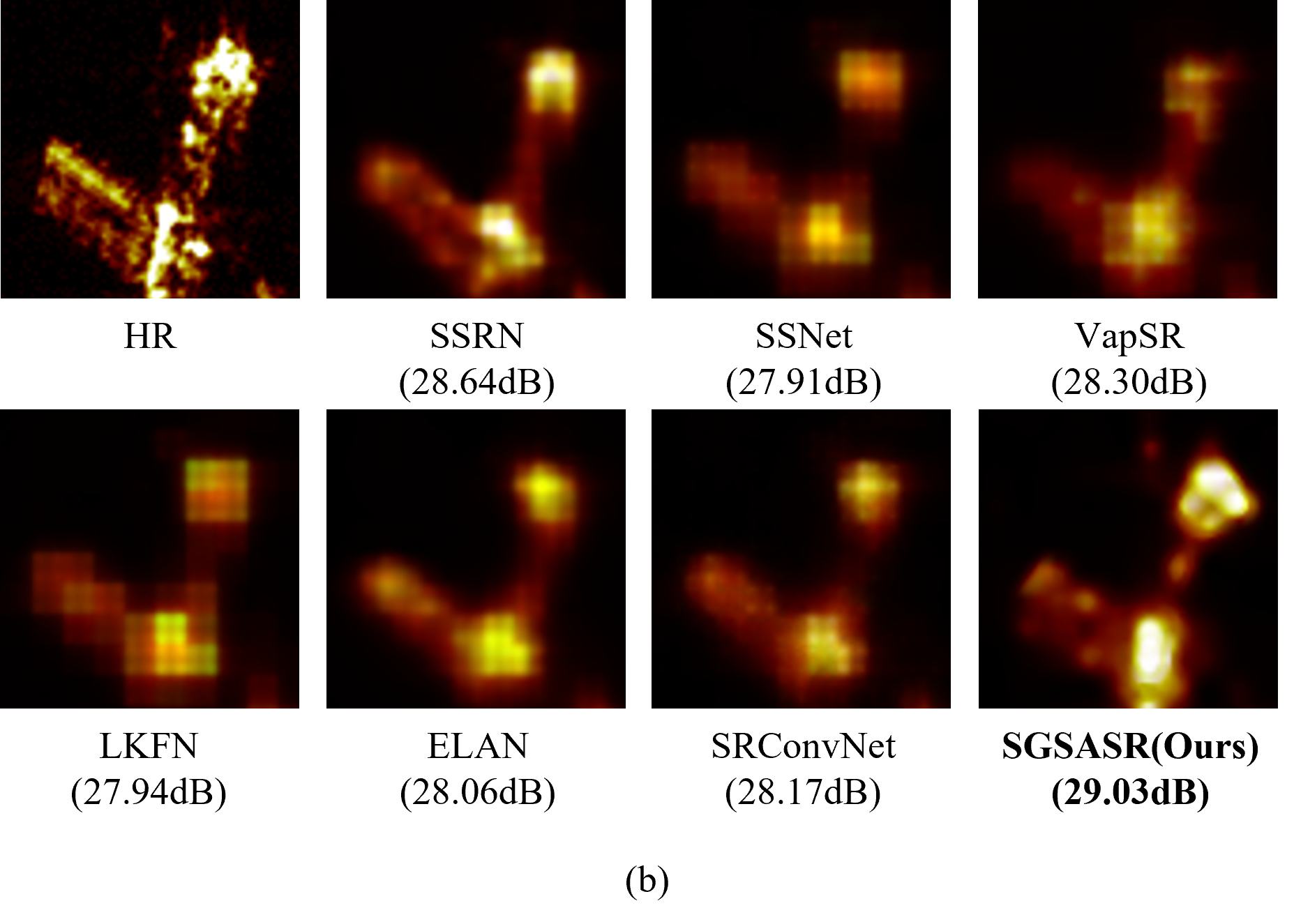}
\end{minipage}  

\caption{Visual comparison for fixed-scale SR methods on spacecraft radar image dataset. (a): $\times4$. (b): $\times8$. }
\label{fig:visual_results_isar_fix}
\end{figure*}

\begin{figure*}[h]
\centering

\begin{minipage}[t]{0.45\linewidth}
  \centering
  \includegraphics[width=1.0\textwidth]{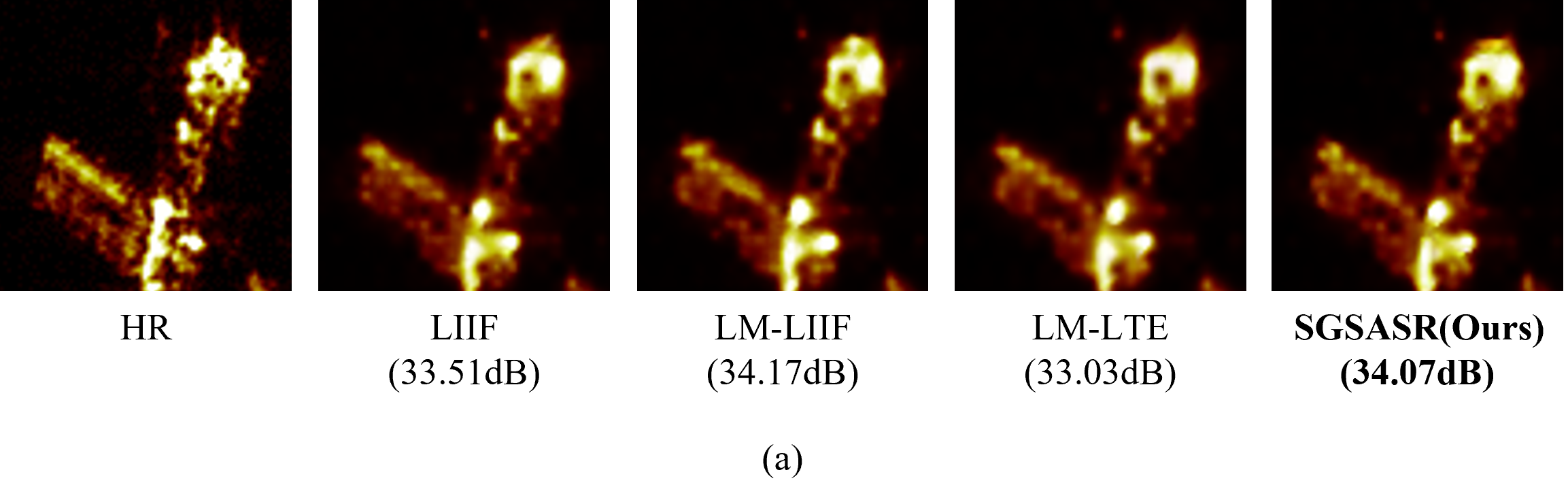}
\end{minipage}  
\quad
\begin{minipage}[t]{0.45\linewidth}
  \centering
  \includegraphics[width=1.0\textwidth]{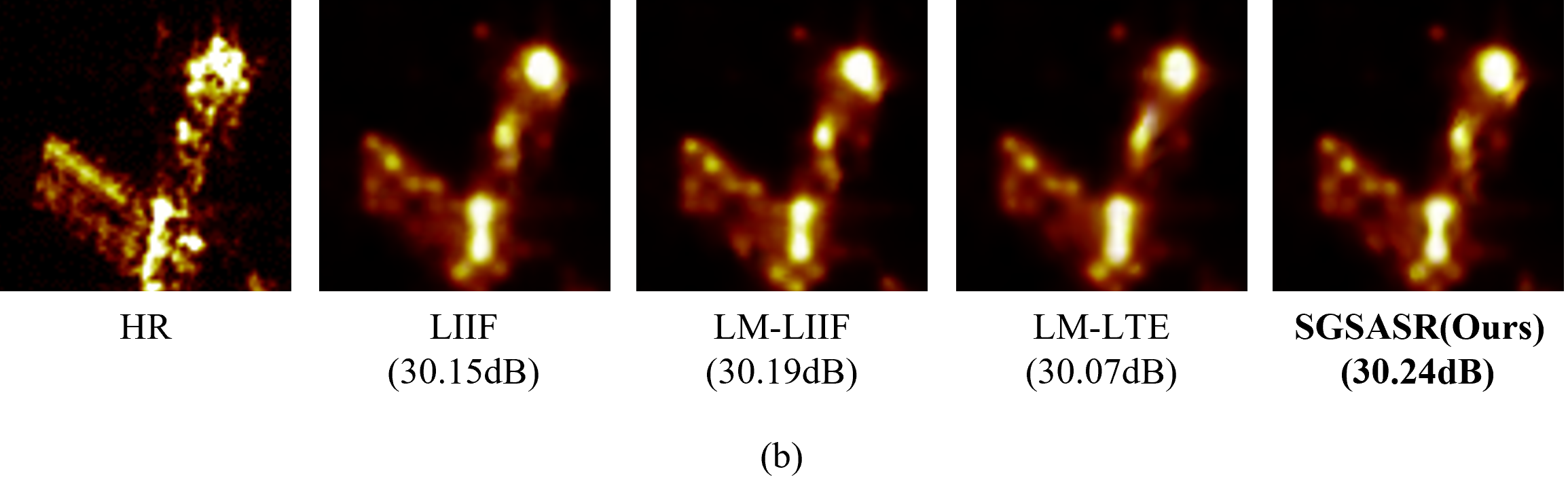}
\end{minipage}  

\begin{minipage}[t]{0.45\linewidth}
  \centering
  \includegraphics[width=1.0\textwidth]{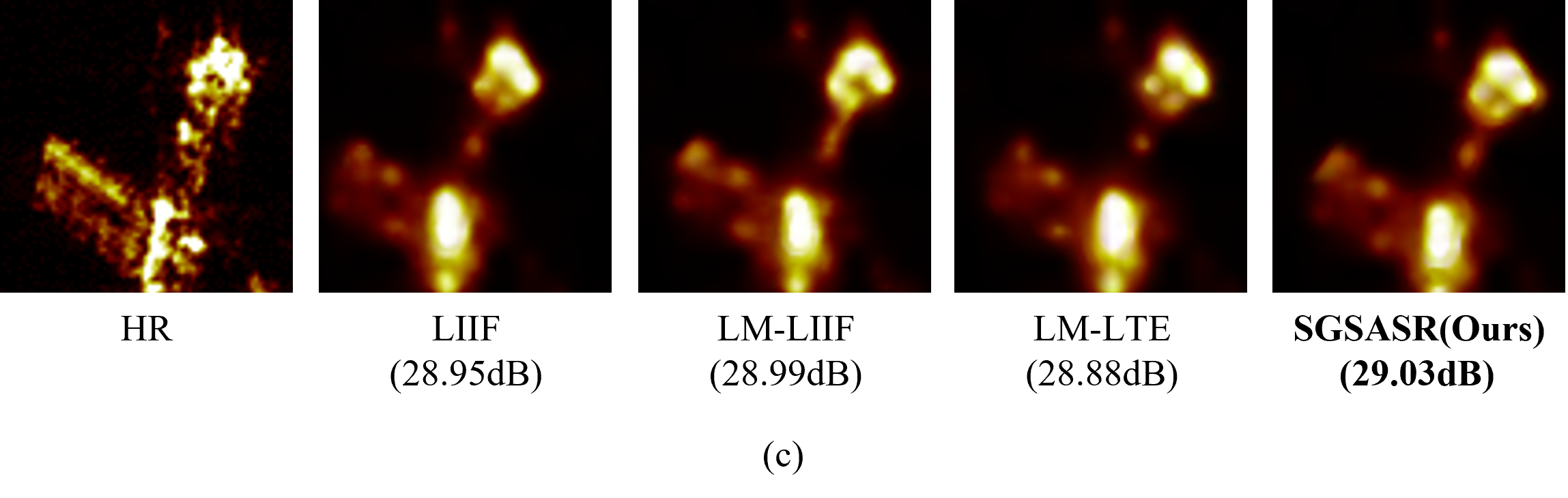}
\end{minipage}  
\quad
\begin{minipage}[t]{0.45\linewidth}
  \centering
  \includegraphics[width=1.0\textwidth]{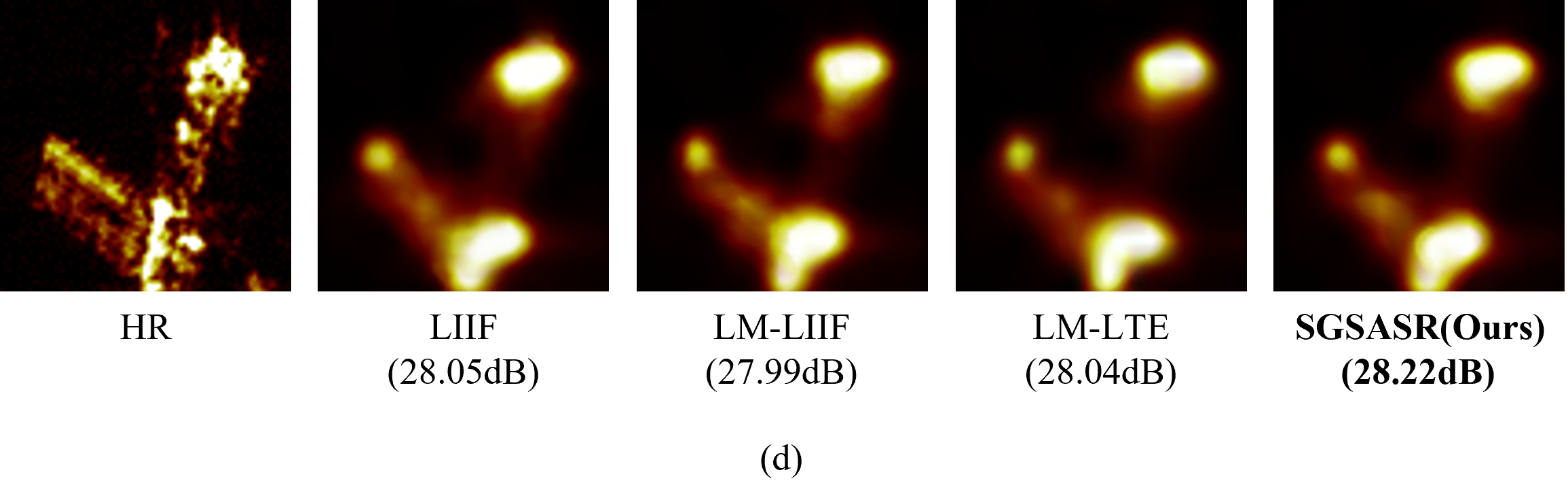}
\end{minipage}

\caption{Visual comparison for arbitrary-scale SR methods on spacecraft radar image dataset. (a): $\times4$. (b): $\times6$. (c): $\times8$. (d): $\times12$.}
\label{fig:visual_results_isar_assr}
\end{figure*}

\subsection{Comparison Results on Spacecraft Radar Image Dataset}
\label{sec:isar_comparison}
In this section, we compare SGSASR with seven fixed-scale SR methods (SSRN~\cite{yang2023ssrn} and SSNet~\cite{gao2024learning}, which are spacecraft image SR methods, as well as VapSR~\cite{zhou2022efficient}, LKFN~\cite{chen2024large}, ELAN~\cite{zhang2022efficient}, MAT~\cite{xie2025mat} and SRConvNet~\cite{li2025srconvnet}), and three arbitrary-scale SR methods (LIIF~\cite{chen2021learning}, LM-LIIF~\cite{he2024latent}, and LM-LTE~\cite{he2024latent}) on the spacecraft radar image dataset. All comparison methods are retrained to ensure fairness and effectiveness. Table \ref{tab:isar_fix_results} presents the quantitative comparison results with fixed-scale SR methods. Table \ref{tab:isar_assr_results} presents the quantitative comparison results with arbitrary-scale SR methods. We can observe that SGSASR significantly outperforms the fixed-scale SR methods in terms of PSNR performance on the spacecraft radar image dataset. Specifically, compared with SSRN, SGSASR achieves improvements of 1.14 dB and 0.39 dB at two SR scales ($\times$4, $\times$8), respectively. We can also observe that SGSASR consistently achieves superior PSNR performance at various SR scales on the spacecraft radar image dataset while maintaining lower computational costs compared with other arbitrary-scale SR methods.

In addition, we present visual comparison results for fixed-scale SR methods in Fig. \ref{fig:visual_results_isar_fix} and for arbitrary-scale SR methods in Fig. \ref{fig:visual_results_isar_assr}. Overall, compared with other fixed-scale SR methods, SGSASR achieves comparable or even better restoration of spacecraft structural details. In contrast to other arbitrary-scale SR methods, the SR images generated by SGSASR not only more effectively restore the structure of the spacecraft, but also avoid introducing noise into the black background region, which clearly demonstrates the superiority of SGSASR.

\begin{figure*}[t]
\centering
\includegraphics[width=1.0\linewidth]{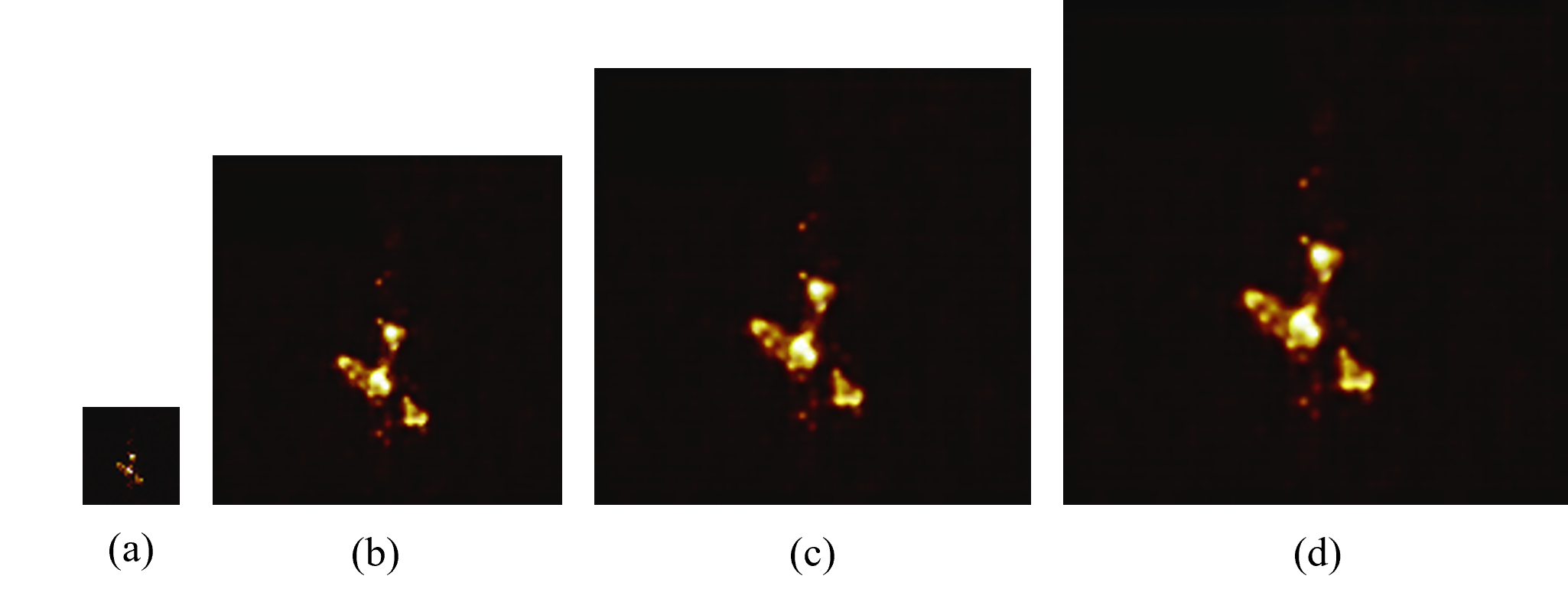}
\caption{Visual results of spacecraft radar image SR using SGSASR at non-integer scales. (a): LR image. (b): $\times$3.6. (c): $\times$4.5. (d): $\times$5.2.}
\label{fig:visual_results_isar_non_integer}
\end{figure*}

Finally, we present visual results of spacecraft radar image SR using SGSASR at three non-integer scales ($\times$3.6, $\times$4.5, $\times$5.2) in Fig. \ref{fig:visual_results_isar_non_integer}. The input spacecraft image is downsampled by a factor of $\times$8 to generate the LR image.

\begin{table*}[t]
\caption{Quantitative comparisons with other fixed-scale SR methods on spacecraft optical image dataset. The best and the second best results are marked in \pmb{bold} and \underline{underline}.}
\label{tab:speed_fix_results}
\centering
\begin{tabular}{lllll}
\hline
\multirow{2}{*}{Methods}& \multicolumn{2}{c}{$\times$4}  & \multicolumn{2}{c}{$\times$8} \\ 
  & \multicolumn{1}{c}{PSNR$\uparrow$}  & \multicolumn{1}{c}{SSIM$\uparrow$} & \multicolumn{1}{c}{PSNR$\uparrow$}  & \multicolumn{1}{c}{SSIM$\uparrow$}\\
\hline
SSRN~\cite{yang2023ssrn} & \underline{33.11} & 0.6967 & \underline{31.99} & \underline{0.6779} \\
SSNet~\cite{gao2024learning} & 32.56 & 0.6607 & 31.30 & 0.6494 \\
VapSR~\cite{zhou2022efficient} & 32.43 & 0.6876 & 30.64 & 0.6501 \\
LKFN~\cite{chen2024large} & 32.96 & \underline{0.7011} & 31.73 & \pmb{0.6813} \\
ELAN~\cite{zhang2022efficient} & 32.83 & 0.6948 & 31.88 & 0.6705 \\
MAT~\cite{xie2025mat} & 32.52 & 0.6891 & 31.00 & 0.6550 \\
SRConvNet~\cite{li2025srconvnet} & 32.95 & 0.6908  & 31.56 & 0.6594 \\
SGSASR(Ours) & \pmb{33.57} & \pmb{0.7132} & \pmb{32.09} & 0.6708 \\
\hline
\end{tabular}
\end{table*}

\begin{table*}[t]
\caption{Quantitative comparisons with other arbitrary-scale SR methods on spacecraft optical image dataset. The best and the second best results are marked in \pmb{bold} and \underline{underline}.}
\label{tab:speed_assr_results}
\centering
\resizebox{\textwidth}{!}{
\begin{tabular}{llllllllll}
\hline
\multirow{2}{*}{Methods}& \multicolumn{2}{c}{$\times$4}  & \multicolumn{2}{c}{$\times$6} & \multicolumn{2}{c}{$\times$8} & \multicolumn{2}{c}{$\times$12} & \multirow{2}{*}{FLOPs$\downarrow$}\\ 
  & \multicolumn{1}{c}{PSNR$\uparrow$}  & \multicolumn{1}{c}{SSIM$\uparrow$} & \multicolumn{1}{c}{PSNR$\uparrow$}  & \multicolumn{1}{c}{SSIM$\uparrow$} & \multicolumn{1}{c}{PSNR$\uparrow$}  & \multicolumn{1}{c}{SSIM$\uparrow$} & \multicolumn{1}{c}{PSNR$\uparrow$}  & \multicolumn{1}{c}{SSIM$\uparrow$}\\
\hline
LIIF~\cite{chen2021learning} & 33.49  & 0.7073 &  32.63 & 0.6772 & 32.01 & 0.6634 & 31.10 & 0.6462 & 3723.64G \\
LM-LIIF~\cite{he2024latent} & 33.44 & 0.7073 & 32.58 & \underline{0.6788} & 31.97 & \underline{0.6647} & 31.07 & \underline{0.6484} & 486.64G \\
LM-LTE~\cite{he2024latent} & \underline{33.54}  & \underline{0.7074} & \underline{32.68} & 0.6764 & \underline{32.06} & 0.6613 & \underline{31.14} & 0.6446 & 499.62G \\
SGSASR(Ours) & \pmb{33.57} & \pmb{0.7132} & \pmb{32.72} & \pmb{0.6851} & \pmb{32.09} & \pmb{0.6708} & \pmb{31.17} & \pmb{0.6531} & \pmb{99.62G}  \\
\hline
\end{tabular}
}
\end{table*}

\begin{figure*}[t]
\centering
\begin{minipage}[t]{0.45\linewidth}
  \centering
  \includegraphics[width=1.0\textwidth]{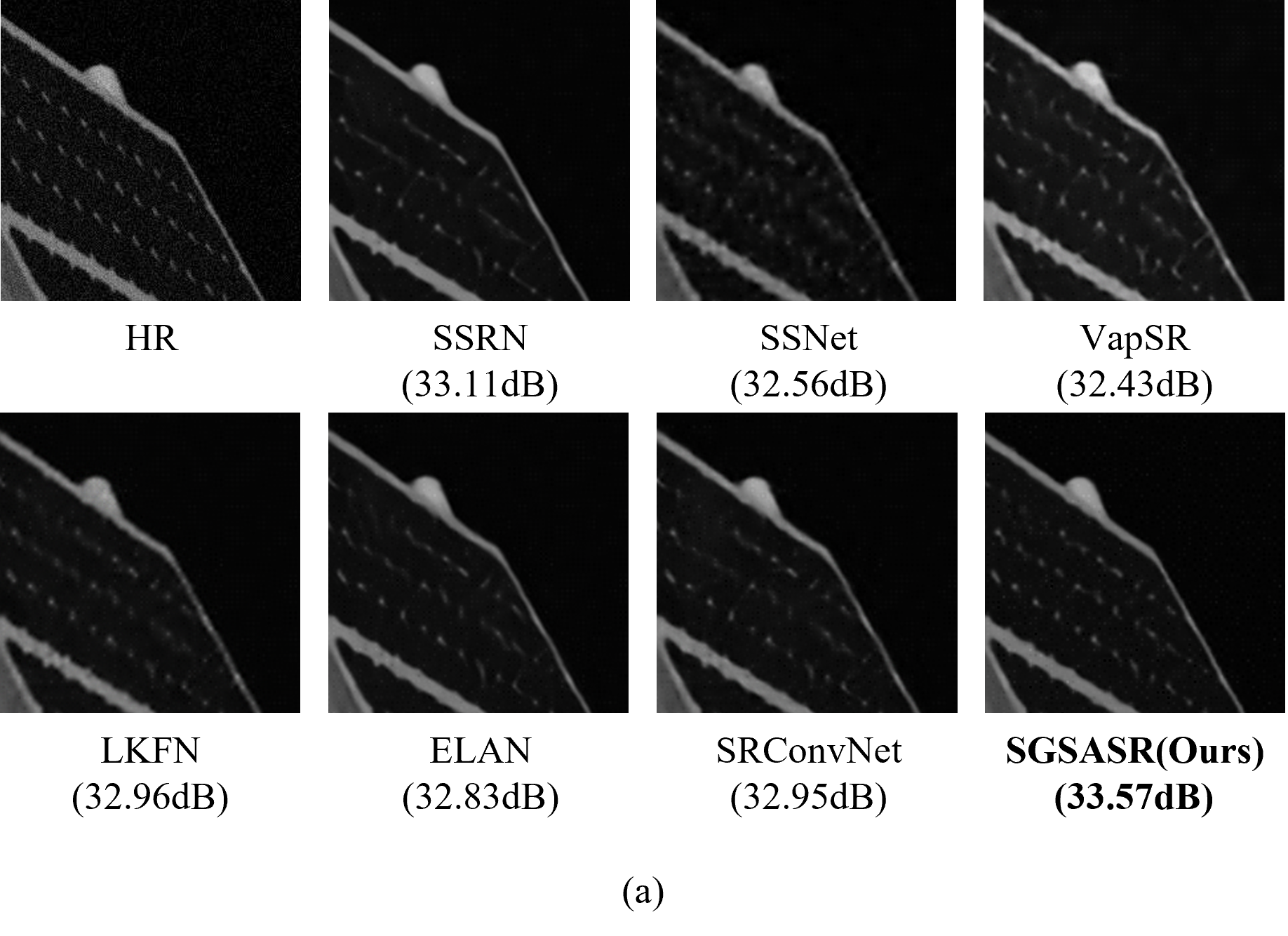}
\end{minipage} 
\quad
\begin{minipage}[t]{0.45\linewidth}
  \centering
  \includegraphics[width=1.0\textwidth]{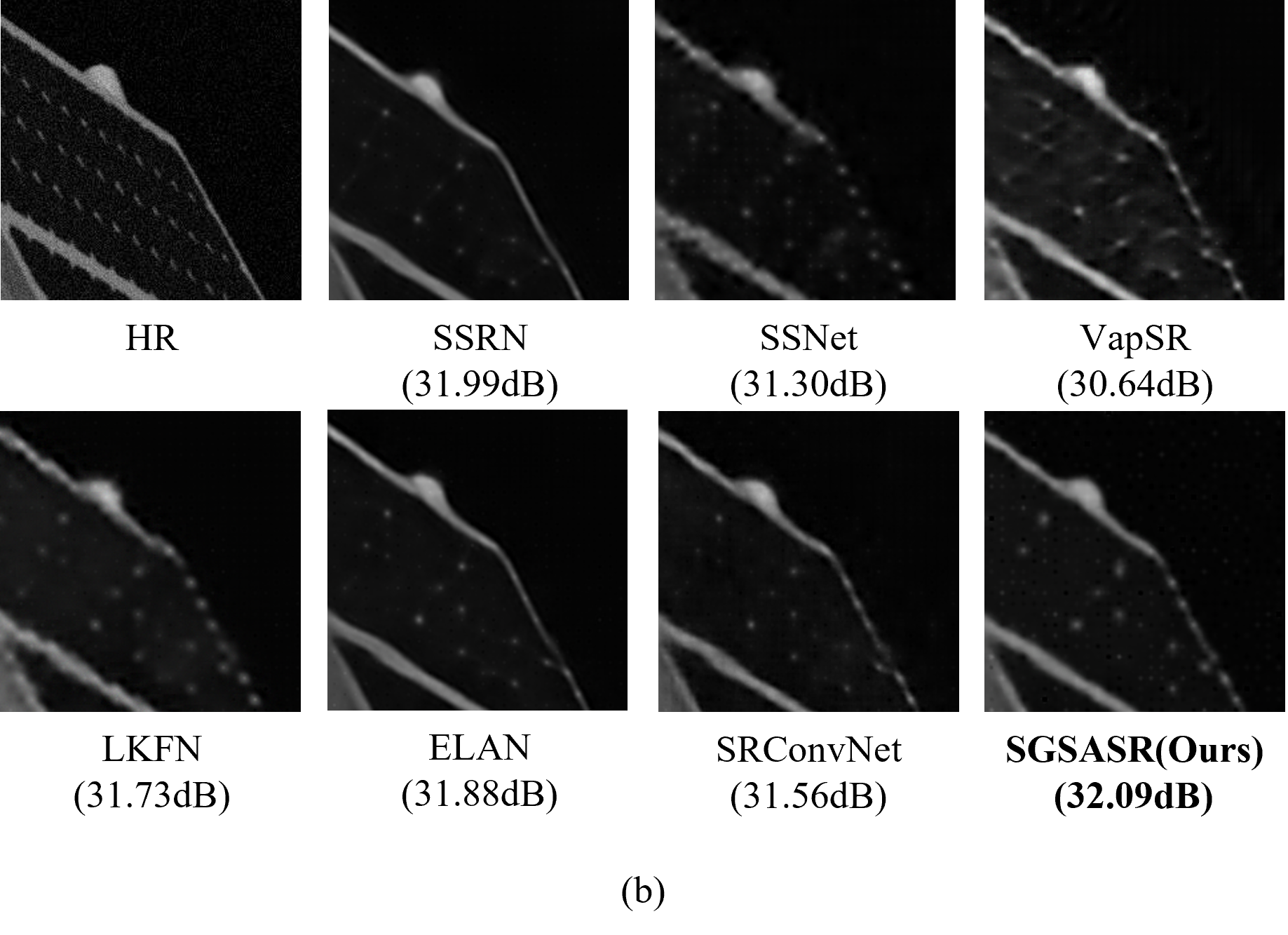}
\end{minipage}  

\caption{Visual comparison for fixed-scale SR methods on spacecraft optical image dataset. (a): $\times4$. (b): $\times8$. }
\label{fig:visual_results_speed_fix}
\end{figure*}

\begin{figure*}[h]
\centering

\begin{minipage}[t]{0.45\linewidth}
  \centering
  \includegraphics[width=1.0\textwidth]{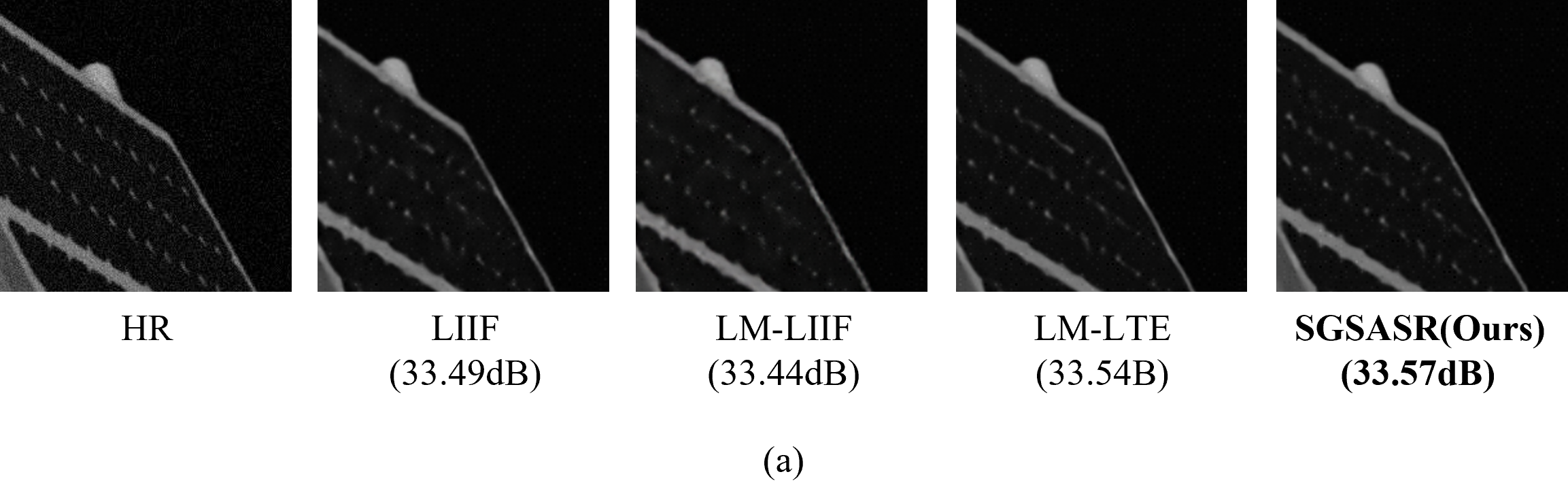}
\end{minipage}  
\quad
\begin{minipage}[t]{0.45\linewidth}
  \centering
  \includegraphics[width=1.0\textwidth]{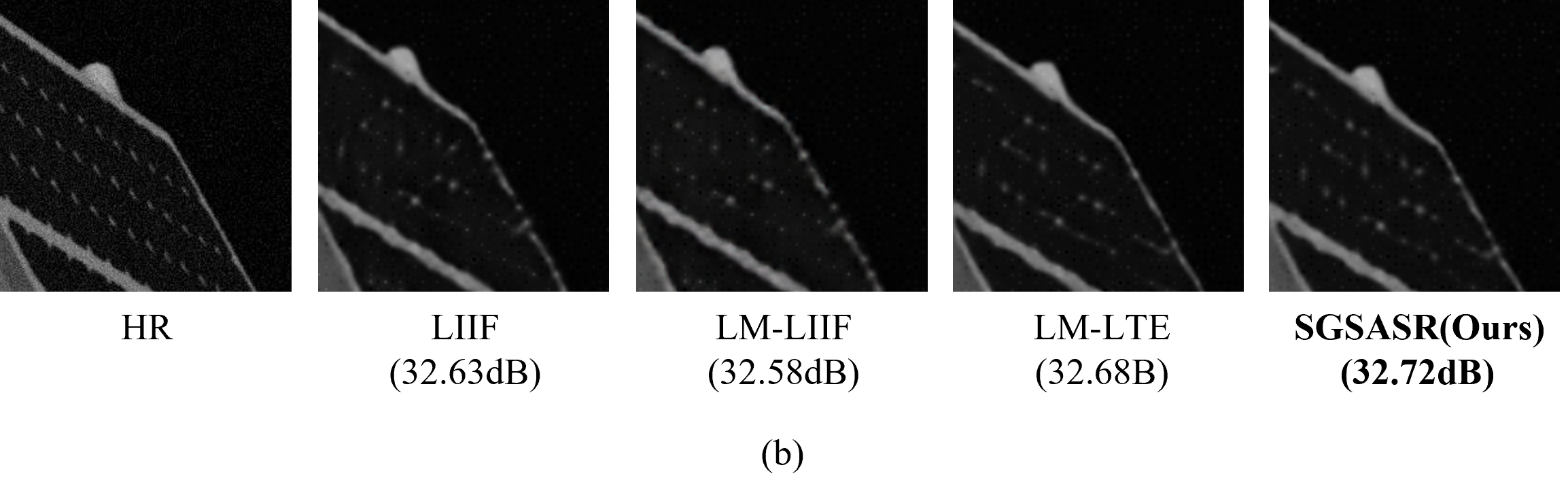}
\end{minipage}  

\begin{minipage}[t]{0.45\linewidth}
  \centering
  \includegraphics[width=1.0\textwidth]{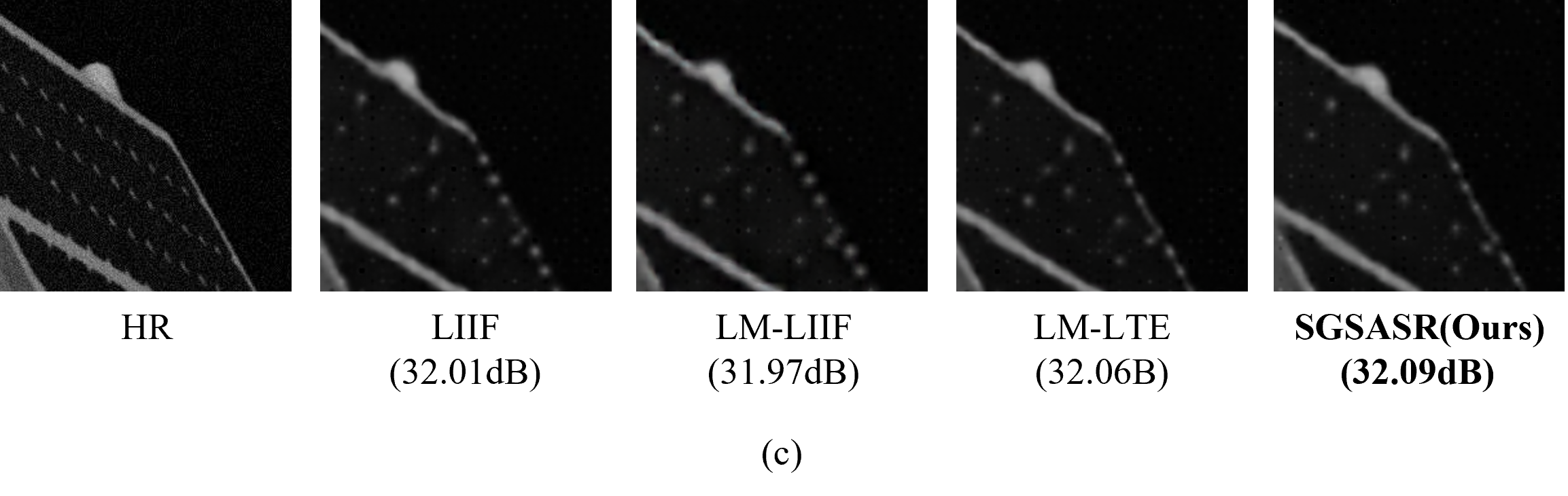}
\end{minipage}  
\quad
\begin{minipage}[t]{0.45\linewidth}
  \centering
  \includegraphics[width=1.0\textwidth]{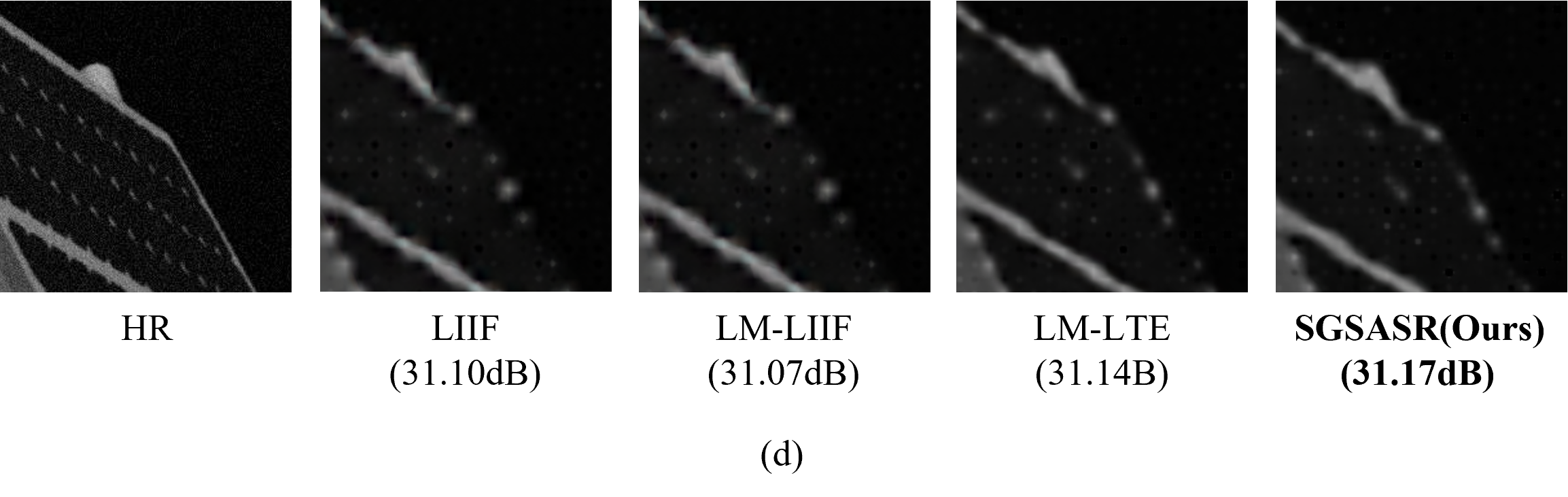}
\end{minipage}

\caption{Visual comparison for arbitrary-scale SR methods on spacecraft optical image dataset. (a): $\times4$. (b): $\times6$. (c): $\times8$. (d): $\times12$.}
\label{fig:visual_results_speed_assr}
\end{figure*}

\subsection{Comparison Results on Spacecraft Optical Image Dataset}
In this section, we compare SGSASR with the ten methods mentioned in Section \ref{sec:isar_comparison} on the spacecraft optical image dataset. All methods are retrained on the spacecraft optical image dataset for a fair comparison. The quantitative comparison results with fixed-scale SR methods are presented in Table \ref{tab:speed_fix_results}. The quantitative comparison results with arbitrary-scale SR methods are presented in Table \ref{tab:speed_assr_results}. We can observe that SGSASR still outperforms fixed-scale SR methods in terms of PSNR performance on the spacecraft optical image dataset. Specifically, compared with SSRN, SGSASR achieves improvements of 0.46 dB and 0.10 dB at two SR scales ($\times$4, $\times$8), respectively. We can also observe that SGSASR consistently achieves superior PSNR performance at various SR scales on the spacecraft optical image dataset while maintaining lower computational costs compared with other arbitrary-scale SR methods.

In addition, we present visual comparison results for fixed-scale SR methods in Fig. \ref{fig:visual_results_speed_fix} and for arbitrary-scale SR methods in Fig. \ref{fig:visual_results_speed_assr}. Overall, compared with other fixed-scale SR methods, SGSASR achieves comparable or even better restoration of spacecraft structural details on the spacecraft optical image dataset. In contrast to other arbitrary-scale SR methods, the spacecraft optical SR images generated by SGSASR not only more effectively restore the structure of the spacecraft, but also avoid introducing noise into the black background regions, which clearly demonstrates the superiority of SGSASR.

\begin{figure*}[t]
\centering
\includegraphics[width=1.0\linewidth]{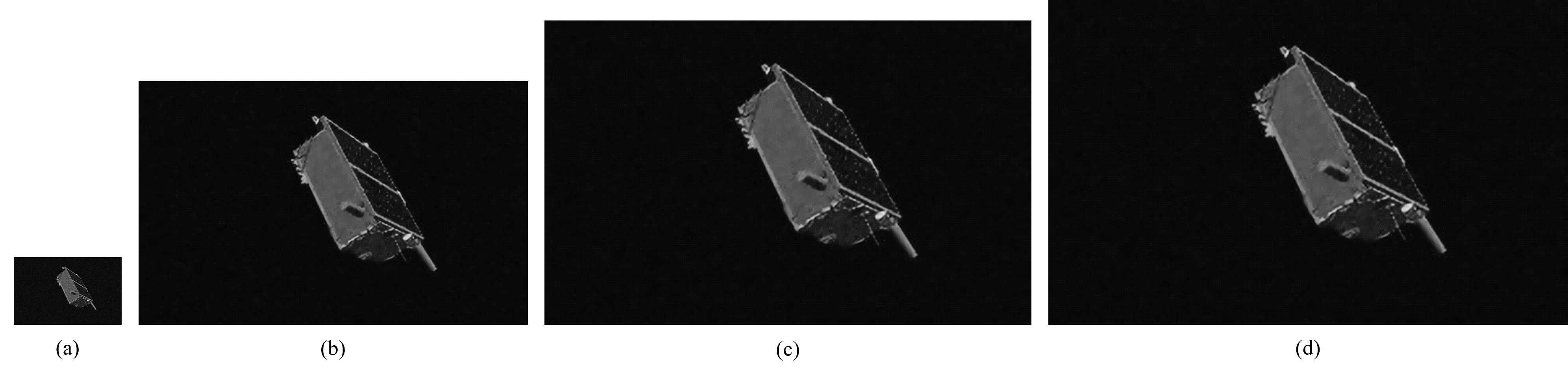}
\caption{Visual results of spacecraft optical image SR using SGSASR at non-integer scales. (a): LR image. (b): $\times$3.6. (c): $\times$4.5. (d): $\times$5.2.}
\label{fig:visual_results_speed_non_integer}
\end{figure*}

Finally, we present visual results of spacecraft optical image SR using SGSASR at three non-integer scales ($\times$3.6, $\times$4.5, $\times$5.2) in Fig. \ref{fig:visual_results_speed_non_integer}. The input spacecraft image is downsampled by a factor of $\times$8 to generate the LR image.

\begin{table}[t]
\caption{Effect of Proposed Modules in SFEEM}
\label{tab:module_ablation}
\centering
\begin{tabular}{c|c|c}
\hline
Methods & PSNR & SSIM \\
\hline
Baseline & 33.72 & 0.9768  \\
Baseline+SCRRB & 33.99 & 0.9820  \\
Baseline+SCRRB+AFFEM & \pmb{34.07} & \pmb{0.9823}\\
\hline
\end{tabular}
\end{table}

\begin{figure}[t]
\centering
\includegraphics[width=1.0\linewidth]{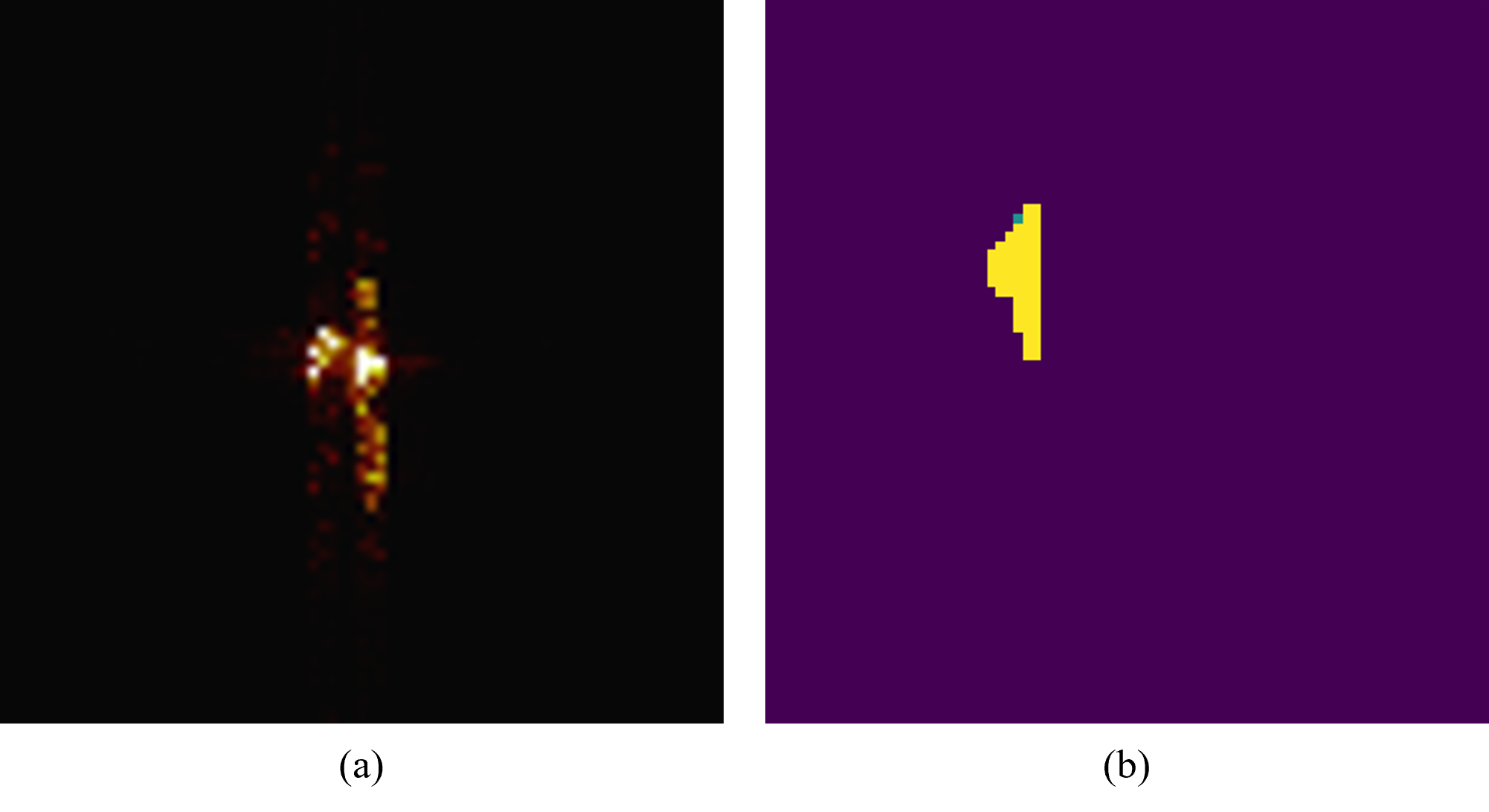}
\caption{Visualization result of SCRRB. (a): The original spacecraft LR image. (b): The feature map extracted by SCRRB. }
\label{fig:vis_result_SCRRB}
\end{figure}

\begin{table}[t]
\caption{Effect of Feature Aggregation Methods}
\label{tab:aggregation_ablation}
\centering
\begin{tabular}{c|c|c}
\hline
Aggregation Methods & PSNR & SSIM \\
\hline
Summation & 33.99 & 0.9820  \\
Concatenation & 33.71 & 0.9816  \\
AFFEM & \pmb{34.07} & \pmb{0.9823} \\
\hline
\end{tabular}
\end{table}

\begin{table}[h]
\caption{Effect of Lantent Modulation in LMASSR}
\label{tab:modulation_ablation}
\centering
\begin{tabular}{c|c|c}
\hline
Lantent Modulation & PSNR & SSIM \\
\hline
$\times$ & 32.27 & 0.9803  \\
Scale & 33.45 & 0.9815  \\
Shift  & 33.79 & 0.9822  \\
Scale+Shift & \pmb{34.07} & \pmb{0.9823} \\
\hline
\end{tabular}
\end{table}

\subsection{Ablation Studies}
\subsubsection{The Effect of Proposed Modules in SFEEM}
To verify the effectiveness of the two main contributions proposed in this paper—--SCRRB and AFFEM, we conduct the incremental ablation experiments on the spacecraft ISAR image dataset. The results are shown in Table \ref{tab:module_ablation}. The first row describes the Baseline model, which directly uses NAFNet~\cite{chen2022simple} as the spacecraft image feature encoder. In the second row, we add the SCRRB, which yields a performance gain of 0.27 dB, since the SCRRB can better identify the core salient region in spacecraft images. In the third row, we further add the AFFEM, which yields an additional performance gain of 0.08 dB, since AFFEM can aggregate the spacecraft region features with general image features more flexibly.

\subsubsection{The Visualization Result of SCRRB}
To demonstrate the recognition ability of our SCRRB for the spacecraft core region, we provide the visualization of the feature map extracted by SCRRB in Fig. \ref{fig:vis_result_SCRRB}. We can observe that the response of the spacecraft body in the feature map is significantly stronger than that of the surrounding black background, indicating that SCRRB can effectively identify the spacecraft core salient region in the image.

\subsubsection{The Effect of Feature Aggregation Methods}
To verify the effectiveness of our AFFEM, we compare it with other feature aggregation methods such as summation and concatenation. The results are shown in Table \ref{tab:aggregation_ablation}. We can observe that AFFEM achieves the best performance, demonstrating its effectiveness in adaptively aggregating the spacecraft region features with general image features.

\subsubsection{The Effect of the Lantent Modulation in LMASSR}
To verify the effectiveness of latent modulation, we compare the performance of different combinations of latent modulation. The results are shown in Table \ref{tab:modulation_ablation}. We can observe that the model using both scale modulation and shift modulation achieves the best performance, indicating that both types of modulation contribute effectively to enhancing arbitrary-scale SR performance.

\section{Conclusion}
\label{sec:conclusion}
This paper introduces SGSASR, an efficient salient region-guided network for spacecraft image arbitrary-scale super-resolution. By leveraging the spacecraft core salient region to guide latent modulation, SGSASR effectively enhances SR performance. In SGSASR, we design a spacecraft core region recognition block (SCRRB) to identify the core salient region in spacecraft images by saliency detection. Additionally, we present an adaptive-weighted feature fusion enhancement mechanism (AFFEM), which flexibly fuses the spacecraft core region features with general image features using dynamic weight parameters. Extensive experiments on multiple datasets demonstrate that our SGSASR not only achieves state-of-the-art performance on spacecraft image arbitrary-scale SR while maintaining high computational efficiency, but also achieves competitive performance compared to fixed-scale SR methods.

This paper mainly studies typical spacecraft images with black space background, and has not yet considered spacecraft images with more complex backgrounds such as the Earth background. Therefore, in future work, we will develop arbitrary-scale super-resolution methods for spacecraft images with more challenging backgrounds.

\section*{CRediT authorship contribution statement}
\pmb{Jingfan Yang:} Conceptualization, Data curation, Investigation, Methodology, Software, Writing – original draft. \pmb{Hu Gao:} Conceptualization, Formal analysis, Writing – review \& editing. \pmb{Ying Zhang:} Investigation, Validation, Conceptualization. \pmb{Depeng Dang:} Funding acquisition, Supervision, Validation, Writing – review \& editing.

\section*{Data availability}
Spacecraft optical image data will be made available on request.

\section*{Acknowledgement}
This work was supported by the National Key Research and Development Program of China (Grant 2020YFC1523303). The authors would like to thank Zhihui Li from China Aerodynamics Research and Development Center for providing the spacecraft radar image dataset.











\bibliographystyle{elsarticle-num} 
\bibliography{sgsasr}

\subsection*{  } 
\noindent \textbf{Jingfan Yang} is currently pursuing the Ph.D. degree
with the School of Artificial Intelligence, Beijing
Normal University, Beijing, China. His research interests include image restoration and image enhancement.\par

\hspace*{\fill} 

\subsection*{  } 
\noindent \textbf{Hu Gao} is currently pursuing the Ph.D. degree
with the School of Artificial Intelligence, Beijing
Normal University, Beijing, China. His research interests include image restoration and image enhancement.\par

\hspace*{\fill} 

\subsection*{  } 
\noindent \textbf{Ying Zhang} is currently pursuing the Ph.D. degree with the School of Artificial Intelligence, Beijing Normal University, Beijing, China. Her research interests include relation extraction and multi modal fusion.\par

\hspace*{\fill} 

\subsection*{  } 
\noindent \textbf{Depeng Dang} receive the Ph.D degree in computer science and technology from the Huazhong University of Science and Technology, Wuhan, China, in 2003. From July 2003 to June 2005, he did his postdoctoral research with the Department of Computer Science and Technology, Tsinghua University, China. Currently, he is a full professor and supervisor of PhD students in computer science from Beijing Normal University, China.\par

\end{document}